\documentclass[twoside]{article}

\usepackage[accepted]{aistats2025}
\usepackage[round]{natbib}

\usepackage[normalem]{ulem}

\newcommand{\camera}[1]{{#1}}

\usepackage{macros} %bbb
\usepackage{amssymb} %bbb
\usepackage{amsmath} %bbb
\usepackage{float} %bbb
\usepackage{graphicx}
\usepackage{enumitem}
\usepackage{hyperref}
\usepackage{cleveref}
\usepackage{mathtools, nccmath, textcomp}
\usepackage[bb=boondox]{mathalfa}
\usepackage{amsmath}
\usepackage{amsfonts}
\usepackage{algorithm}
\usepackage{algorithmic}
\usepackage{mdframed}
\newmdenv[linecolor=blue,backgroundcolor=blue!10,innerleftmargin=5pt,innerrightmargin=5pt,innertopmargin=2pt,innerbottommargin=2pt]{bluebox}

\usepackage{soul}  % for highlighting
\soulregister{\ref}{1}  % Register \ref
\soulregister{\~}{1}    % Register \~

\sethlcolor{blue!10}
\usepackage{multirow}
\usepackage{graphicx}
\usepackage[table,xcdraw]{xcolor}

\begin{document}



\twocolumn[

\aistatstitle{Federated Communication-Efficient Multi-Objective Optimization}

\aistatsauthor{ Baris Askin \And Pranay Sharma \And  Gauri Joshi \And Carlee Joe-Wong }

\aistatsaddress{ Carnegie Mellon University} ]

\begin{abstract}
We study a federated version of multi-objective optimization (MOO), where a single model is trained to optimize multiple objective functions. MOO has been extensively studied in the centralized setting but is less explored in federated or distributed settings. We propose $\our$, a novel communication-efficient federated multi-objective optimization (FMOO) algorithm that improves the error convergence performance of the model compared to existing approaches. Unlike prior works, the communication cost of $\our$ does not scale with the number of objectives, as each client sends a single aggregated gradient to the central server. We provide a convergence analysis of the proposed method for smooth and non-convex objective functions under milder assumptions than in prior work. In addition, we introduce a variant of $\our$  that allows users to specify a preference over the objectives in terms of a desired ratio of the final objective values. Through extensive experiments, we demonstrate the superiority of our proposed method over baseline approaches.
\end{abstract}

\section{INTRODUCTION}

Multi-objective optimization (MOO) has gained significant attention lately due to its role in the success of multi-task learning (MTL) \citep{caruana1997multitask, evgeniou2004regularized, zhang2021survey}, especially in practical applications such as computer vision \citep{mtl_cv_facial_landmark}, natural language processing \citep{mtl_nlp}, reinforcement learning \citep{mtl_rl}, experimental design \citep{christensen2021data}, and power systems \citep{yin2021analytical}. In MTL, the data from multiple related tasks is used to simultaneously learn model(s) for the tasks. MOO is an instance of MTL where the goal is to train a single model $\x$ that simultaneously optimizes multiple objectives:
\begin{align}
    &\min_{\x \in \mathbb{R}^\xdim} \lvect \p{\x} := \sqbr{\tl_1(\x), \tl_2(\x), \dots, \tl_\M(\x)}^\top,
\label{eq:moo_general_problem}
\end{align}
where $\M$ is the number of objectives, $\lvect \in \mathbb{R}^\M$ is the vector of the $\M$ individual objective loss functions $\{\tl_i \}_{i=1}^\M$, and $\x\in\mathbb{R}^\xdim$ is the common model that seeks to minimize them. In general, there need not exist a model $\x$ that simultaneously minimizes all the $M$ losses. On the contrary, minimizing one loss often adversely affects performance on other tasks \citep{kurin2022defense}. This phenomenon is commonly referred to as \textit{negative transfer} \citep{mueller2024can, zhang2022survey}. Therefore, MOO algorithms instead aim to find a \textit{Pareto optimal/stationary} solution, where any further improvement in one loss necessarily worsens another (see definitions in \Cref{sec:prelim}) 
\citep{senermtl}. This results in a trade-off across the objectives, for example, balancing treatment efficacy and side effects in personalized medicine \citep{drug_efficacy_side_effect}. Preference-based MOO \citep{epo} extends traditional MOO by incorporating the user preferences of relative task priorities into the model training. This enables users to tailor the solution based on their preferences, rather than seeking an arbitrary Pareto solution.

With the increasing proliferation of mobile and Internet-of-Things devices, an ever-increasing proportion of data for many applications, including many MOO problems,
comes from edge devices. Federated learning (FL) \cite{fedavg, kairouz2021advances} is a distributed learning paradigm that facilitates efficient collaborative learning of machine learning models at the edge devices, or \textit{clients}. In FL, clients train the models using their local data. With the ever-growing size of machine learning models \citep{villalobos2022machine}, communication with the central server is the main bottleneck. Therefore, the local models are only periodically aggregated at the server. Existing work addresses the multiple challenges faced by FL algorithms, including fairness, privacy, and robustness to heterogeneity \citep{cui2021addressing, hu2022federated, mehrabi2022towards}. 
However, the proposed solutions usually focus on solving a single-objective optimization problem, leaving federated multi-objective optimization relatively underexplored. 

This work focuses on the federated multi-objective optimization (FMOO) problem. In addition to the healthcare applications mentioned earlier, FMOO applications include recommendation systems \citep{sun2022survey}, where relevance, product ratings, and personal preferences must be optimized simultaneously using private user data. 
In both of these applications, federated approaches are essential to protect the privacy of user preferences and health records. 
Compared to centralized MOO, FMOO faces additional challenges, including data heterogeneity across clients and excessive communication requirements, making learning more difficult. For example, the communication cost in \citep{FMGDA}, one of the few existing works on FMOO, scales linearly with the number of objectives, $\M$. In our work, we address these challenges.

\paragraph{Contributions.} We consider a multi-objective optimization problem in the federated setting. After reviewing some related work in \Cref{sec:related_work} and preliminaries in \Cref{sec:prelim}, we make the following contributions:
\begin{itemize}[noitemsep, leftmargin=*]
\setlength\itemsep{0.1em}
    \item We propose $\our$\footnote{The code is provided at \href{https://github.com/askinb/FedCMOO}{https://github.com/askinb/FedCMOO}.}, a communication-efficient federated algorithm to solve the MOO problem (\ref{eq:moo_general_problem}). Unlike existing work, the communication complexity of $\our$ does not scale with the number of objectives $\M$ (see \Cref{sec:algo}). 
    \item We theoretically prove the convergence of $\our$ for smooth non-convex objectives under milder assumptions than prior work (\Cref{thm:FedCMO}). Also, the sample complexity of $\our$ has a better dependence on the number of objectives $\M$. (\Cref{sect:theory}).
    \item We propose $\ourp$, to our knowledge, the first federated 
    algorithm to train a model that satisfies user-specified preferences, in terms of balancing the different objective values (\Cref{sect:preference_alg}).
    \item Through extensive experiments, we demonstrate the superior performance and efficiency of our proposed methods (\Cref{sec:experiments}).
\end{itemize}

\section{RELATED WORK}
\label{sec:related_work}

\paragraph{Multi-objective Optimization.} Existing approaches to solve multi-objective optimization problems include gradient-free methods, such as evolutionary methods \citep{deb2002fast, vargas2015general} and Bayesian optimization \citep{belakaria2020uncertainty}. However, gradient-based methods \citep{mgda, fliege2019complexity, liu2024stochastic, peitz2018gradient} are more suited to solving high-dimensional problems \citep{senermtl} and form the focus of our work. Following the seminal work of \cite{mgda}, several approaches to MOO have been proposed in the literature. The simplest of these is \textit{linear scalarization} \citep{hu2024revisiting, lin_smooth_2024, liu2021conflict, royer2024scalarization}, where the $M$-dimensional vector of losses in (\ref{eq:moo_general_problem}) is replaced by a convex combination of the losses, resulting in a simpler scalar minimization. However, this approach suffers in the presence of \textit{conflicting} gradients across tasks \citep{yu2020gradient}. In addition, \cite{hu2024revisiting} show that linear scalarization fails to explore the entire Pareto front. Several works propose algorithms to address the problem of conflicting gradients in MOO in the centralized setting \citep{kurin2022defense, royer2024scalarization, xin2022current}.

\paragraph{Federated MOO and MTL.} In our work, we consider the MOO problem in a federated setting where the goal is to train a model to simultaneously minimize multiple objective functions whose data is distributed across clients. Recently, some works have started exploring this setting in the decentralized \citep{blondin2021decentralized} and federated setting \citep{FMGDA}. However, their proposed algorithms incur high communication overhead, and the accompanying analyses make stronger assumptions than our work, as we explain in more detail in later sections. Another related, but distinct line of work is on multi-task learning for the purpose of personalization in federated learning \citep{smith2017federated, chen2023fedbone, lu2024fedhca2, marfoq2021federated, mills2021multi}. These works still consider one task at all clients (e.g., an image classification problem) and aim to train a personalized model for each client, or train models to optimize different metrics (e.g., accuracy, fairness, or robustness against adversaries) \citep{cui2021addressing, mehrabi2022towards, hu2022federated}.

\paragraph{Preference-based MOO.}
Pareto optimal solutions with specific trade-offs between losses are often desirable. Some applications include recommender systems \citep{milojkovic2019multi}, drug design \citep{jain2023multi}, image classification \citep{raychaudhuri2022controllable}, and reinforcement learning \citep{basaklar2022pd}. \citep{epo} proposed the first gradient-based multi-objective method to reach a preference-based solution, followed by more recent work in 
\citep{ruchte2021scalable, zhang2024pmgda}. A related line of work aims to discover the entire Pareto set, rather than a single solution with a preference \citep{chen2022multi, dimitriadis2023pareto, haishan2024preference, lin2022pareto}. In our work, we extend the preference-based algorithm of \cite{epo} to a federated setting.

\nocite{lin2024few}

%\section{PROBLEM \& METHOD}
\section{PRELIMINARIES}
\label{sec:prelim}
%\subsection{Preliminaries}
\paragraph{Notations.}
Given a positive integer $\kk$, we define the set $[\kk] \triangleq \{1,\dots,\kk\}$. $\G$ and $\sG$ denote the gradient and stochastic gradient operators respectively. When used with a vector function, these operators return the Jacobian matrix, e.g., \mbox{$\sG\lvect\p{\x} = [\sG \tl_1(\x), \dots, \sG \tl_\M(\x)] \in \mathbb{R}^{\xdim \times \M}$}. We use bold letters (e.g., $\x$) to denote vectors. $|S|$ is the cardinality of the set $S$. $\|\cdot\|$ denotes the Euclidean norm.

\paragraph{Pareto Optimal and Stationary Solutions.} 
Unlike single-objective optimization, with MOO, which aims to solve the problem in \ref{eq:moo_general_problem}, all the objectives generally cannot be optimized simultaneously. The inherent trade-offs between objectives lead to non-dominated solutions, where any further improvement in one objective necessarily implies making another objective worse. A solution $\x^*$ is called \textit{Pareto optimal} if it is non-dominated, i.e., there exists no other $\x\in \mathbb{R}^\xdim$ such that $\tl_\kk\p{\x}\leq\tl_\kk\p{\x^*}$ for all $\kk\in[\M]$ and $\tl_{\kk'}\p{\x}<\tl_{\kk'}\p{\x^*}$ for some $\kk'\in[\M]$. The set of (potentially infinite) Pareto optimal solutions is called Pareto front, with each point representing a different trade-off among the $M$ objectives. However, for non-convex problems, finding a Pareto optimal solution is NP-hard in general. The solution $\x^*$ is called \textit{Pareto stationary} if there exists $\w\in\WM$ such that 
$\sum_\kk\wk\G\tl_k=\mathbf{0}$, where $\WM$ denotes the probability simplex, i.e., zero vector is in the convex hull of the gradients of objectives. At Pareto stationary points, there is no common descent direction for all objective functions. Pareto optimal solutions are also Pareto stationary.

\paragraph{Multi-gradient descent algorithm ($\MGDAnormal$).}  
Among iterative gradient-based methods, the multi-gradient descent algorithm (\(\MGDA\)) \citep{mgda, senermtl, direction_oriented} guarantees convergence to \textit{a Pareto stationary solution}. In $\MGDA$, given some small learning rate $\lr$, the descent direction maximizes the minimum descent across tasks, i.e.,
\begin{align}
\label{eq:max-min-descent}
    \descent^* = \max_{\descent\in\mathbb{R}^\xdim} \min_{\kk\in[\M]} \tl_\kk\p{\x} - \tl_\kk\p{\x - \lr \descent},
\end{align}
where $\x$ is the current model. Using first-order Taylor approximation, this optimization problem can equivalently be solved by finding the optimal weights for a convex combination of the objective gradients:
\begin{align}
    \w^*= \argmin_{\w \in \WM} \norm{\textstyle\sum_k \wk\nabla\tl_\kk\p{\x}}, \label{eq:mgda_obj}
\end{align}
where $\WM$ is the probability simplex in $\mathbb{R}^{\M}$.
The descent vector is then \mbox{$\descent^* = \sum_k \wk^* \nabla \tl_\kk\p{\x}$.}
Please refer to \Cref{app_sect:moo_and_mgda} for more details.

For large-scale applications, computing exact gradients is computationally expensive. Therefore, \texttt{Stochastic-}$\MGDA$ \citep{liu2021conflict, zhou2022convergence} uses stochastic gradients in (\ref{eq:mgda_obj}). However, stochasticity also introduces additional theoretical challenges, and recent work focuses on proposing provably convergent stochastic MOO algorithms \citep{three_way_tradeoff, fernando2022mitigating, direction_oriented}.

\paragraph{Federated MOO.}
We consider the federated MOO problem (\ref{eq:moo_general_problem}), with $\N$ clients and a central server. For simplicity, we assume that each client has data for all tasks. However, our results easily extend to the more general case, where each client has data corresponding to only a subset of the $M$ objectives. 
We denote the loss function corresponding to task $\kk$ at client $\ii$ as $\fik$, and the global loss function for task $\kk$ is defined as:
% \sout{the average of the local objectives:}
\begin{align}
    \tl_\kk\p{\x} = \frac{1}{\N} \sum_{\ii=1}^\N \fik\p{\x}, \quad \forall \kk \in [\M].
    \label{eq:lossTask}
\end{align}
When solving (\ref{eq:moo_general_problem}) in FL, the server communicates the model $\x$ to the clients. On the other hand, clients retain their data and only communicate their gradients or accumulated local updates to the server. 

Closest to our work, \citep{FMGDA} introduces Federated Stochastic MGDA ($\fedmgda$). In one round of $\fedmgda$, each participating client performs local updates separately for all $\M$ objectives, and then sends the $\M$ individual updates to the server. The server averages these $\M$ updates across clients and uses them to compute the aggregation weights (\ref{eq:mgda_obj}). 
% Please refer to \Cref{app_sect:moo_and_mgda} for further details. 
As a result, $\fedmgda$ faces two issues: 1) \textit{High communication cost}, scaling linearly with the number of objectives $\M$, since clients upload $\xdim$-dimensional updates for each objective; and 2) \textit{Significant local drift across objectives}, due to separate local trainings at clients, which deteriorates the performance of $\fedmgda$ (see \Cref{sec:experiments}).

Next, we discuss our proposed algorithm $\our$ (\underline{Fed}erated \underline{C}ommunication-efficient \underline{M}ulti-\underline{O}bjective \underline{O}ptimization) that addresses these shortcomings. 

\section{PROPOSED 
% \texorpdfstring{\(\ournormal\)}{FedCMOO} 
ALGORITHM}
\label{sec:algo}
Our proposed $\our$ (\Cref{alg:our}) follows the same idea as \(\MGDA\). Once the server computes aggregation weights, the clients perform local training using the weighted sum of objectives and send updates to the server. The server then aggregates the client updates to compute the descent direction.

\setlength{\algorithmicindent}{.52em}  % indentation level
\begin{algorithm}
\caption{\our}
\label{alg:our}
\begin{algorithmic}[1]
\STATE \textbf{Input:} client and server learning rates $\lrl, \lrg$, number of local steps $\locit$
\STATE \textbf{Initialize:} global model $\x^\tp{0} \in \mathbb{R}^\xdim$, and task weights $\w^\tp{0}\gets [1/\M,\dots,1/\M]^\top \in \WM$\label{line:our_init}
\FOR{$t = 0, 1, \dots, T - 1$}
    \STATE Select the client set $\B^\tp{t}$ unif. randomly from $[N]$; send $\x^\tp{t}$ to the clients in $\B^\tp{t}$ \label{line:our_client_selection}
    \STATE $G^\tp{t} \gets \approxg (\B^\tp{t}, \x^\tp{t})$ at the server, where $G^{(t)} \approx \G \lvect (\x^{(t)})^\top \G \lvect (\x^{(t)})$ \label{line:jacobian_approx}
    \STATE Compute $\w^\tp{t+1} \gets \getw$($\w^\tp{t}, G^\tp{t}$) at server and send to clients in $\B^\tp{t}$ \label{line:our_wupdate}
    \FOR{each client $\ii \in \B^\tp{t}$ \textbf{in parallel}}
        \STATE Initialize local model: $\x_\ii^\tp{t,0} \gets \x^\tp{t}$ \label{line:client_update_begin}
        \FOR{$r = 0, \dots, \locit-1$}
            \STATE \label{line:our_local_sgd}
            \(
            \x_\ii^\tp{t,r+1} \gets \x_\ii^\tp{t,r} - \lrl \sum_{k=1}^\M \wk^\tp{t+1} \sG\fik (\x_\ii^\tp{t,r})
            \) 
        \ENDFOR \label{line:client_update_end}
        \STATE Send
        \(
        \D_\ii^\tp{t} \triangleq \mfrac{1}{\locit \lrl} \big( \x^\tp{t} - \x_\ii^\tp{t,\locit} \big)
        \) to the server \label{line:our_update_calc}
    \ENDFOR
    \STATE
    \(
    \x^\tp{t+1} \gets \x^\tp{t} - \lrg \lrl \locit \mfrac{1}{|\B^\tp{t}|} \sum_{\ii \in \B^\tp{t}} \D_\ii^\tp{t}
    \) \label{line:our_aggregation}
\ENDFOR
\STATE \textbf{Return} $\x^\tp{T}$ 
\end{algorithmic}
\end{algorithm}

At the beginning of each round $t$, the server selects clients uniformly at random and sends the current model $\x^\tp{t}$ to them (\Cref{alg:our}, line~\ref{line:our_client_selection}). Next, the server approximates the Gram matrix of the Jacobian of loss vector $G^{(t)} \approx \G\lvect (\x^\tp{t})^\top \G\lvect (\x^\tp{t})$, with $\approxg$ subroutine (line \ref{line:jacobian_approx}). For this approximation, each client in $\B^\tp{t}$ needs to calculate a stochastic Jacobian matrix and send it to the server in a communication-efficient way, as described in \Cref{alg:approxg} in detail. The server uses $G^{(t)}$ to calculate the new aggregation weights $\w^\tp{t+1}$ with the $\getw$ subroutine (\Cref{alg:getw_G}) and sends these weights to the selected clients (line~\ref{line:our_wupdate}).
% The $\approxg$ (\Cref{alg:approxg}) and $\getw$ (\Cref{alg:getw_G}) subroutines are explained in more detail below. 
Next, each participating client $i$ performs $\locit$ local SGD steps on the weighted aggregate loss $\sum_{k=1}^\M \wk^\tp{t+1} \fik$ (lines \ref{line:client_update_begin}-\ref{line:client_update_end}). When finished, the clients send their updates $\{\D_\ii^{(t)}\}_{i \in \B^{(t)}}$ to the server, which aggregates these to arrive at the updated model $\x^{(t+1)}$, concluding one round of $\our$.

Next, we discuss the two subroutines used in $\our$, $\getw$ and $\approxg$. 

\paragraph{$\getwnormal$.}
As the name suggests, $\getw$ computes the aggregation weights $\{\w^{(t)}\}$ needed to compute the descent direction (\ref{eq:mgda_obj}). The server solves the convex minimization in (\ref{eq:mgda_obj}) using an iterative method, such as projected gradient descent (PGD). A single PGD update takes the form
\begin{equation}
    \w \gets \projwm \p{\w-\beta\G\lvect\p{\x}^\top\G\lvect\p{\x} \w},
    \label{eq:findWeight}
\end{equation}
where $\beta$ is the step size. The server can also utilize an approximation $G$ of $\G \lvect \p{\x}^\top \G \lvect\p{\x}$ in (\ref{eq:findWeight}), resulting in \Cref{alg:getw_G}, which runs $K$ PGD steps. For $\M \ll \xdim$, the computational cost of $\getw$ is negligible \camera{since $G\in\mathbb{R}^{\M\times\M}$ and $\w\in\mathbb{R}^{\M}$}.
 
\setlength{\algorithmicindent}{.52em}  % indentation level
\begin{algorithm}
\caption{$\getw$}
\label{alg:getw_G}
\begin{algorithmic}[1]
\STATE \textbf{Input:} weights $\w$, Gram matrix $G$, step-size $\beta$, number of iterations $K$
% \STATE \ps{Can omit time index $t$, by adding $\w$ as input}
% \STATE $\w^\tp{t} \gets \projwm\p{\w^\tp{t}-\lrb G\w^\tp{t}}$ for $K$ iterations \label{line:w_update}
% \STATE \textbf{Return} $\w^\tp{t}$
\STATE $\w \gets \projwm\p{\w-\lrb G\w}$ for $K$ iterations \label{line:w_update}
\STATE \textbf{Return} $\w$
\end{algorithmic}
\end{algorithm}

Next, we discuss the construction of \mbox{$G \approx \G \lvect \p{\x}^\top \G \lvect\p{\x}$} at the server.

\paragraph{Randomized SVD-based Jacobian Approximation.}
Naively constructing $\G\lvect\p{\x} = \sqbr{\G \tl_1(\x), \G \tl_2(\x), \dots, \G \tl_\M(\x)} \in \mathbb{R}^{\xdim \times \M}$ at the server would require each participating client $i$ to \camera{calculate and} communicate its gradients $\{ \G \fik (\x)\}_{k=1}^{\M}$. Even constructing the stochastic version $\widetilde{\G} \lvect\p{\x}$ with stochastic gradients
would incur a communication cost of $\Theta({\M\xdim})$ per client, which scales linearly with the number of objectives $\M$. In the $\approxg$ subroutine, we propose a randomized SVD-based \citep{randsvd} approach to approximate the Gram matrix \mbox{$\G \lvect \p{\x}^\top \G \lvect \p{\x}$}, which reduces the per-client communication cost from $\Theta(\M \xdim)$ to $\Theta(\xdim)$. 
\camera{We note that the proposed $\our$ framework is compatible with any compressed communication method, which can be used as a substitute for the Randomized SVD-based $\approxg$. To effectively capture the correlation across task gradients, we specifically design $\approxg$ and empirically validate its performance against several widely used compression techniques in Appendix~\ref{app_sect:approxg_details}.}

\setlength{\algorithmicindent}{.52em}  % indentation level
\begin{algorithm}
\caption{$\approxg$}
\label{alg:approxg}
\begin{algorithmic}[1]
\STATE \textbf{Input:} participating client set $\B$, model $\x$
\FOR{each client $\ii \in \B$ in parallel\label{line:for_loop_clients}}
    \STATE $\hh_\ii \triangleq \sqbr{\sG\f_{\ii,1}\p{\x},\dots,\sG\f_{\ii,\M}\p{\x}} \in \mathbb{R}^{\xdim \times \M}$\label{line:hi_calculation}
    \STATE $\hs_\ii \gets \text{Reshape}(\hh_\ii)$ to $\mathbb{R}^{\sqrt{\xdim\M} \times \sqrt{\xdim\M}}$ \label{line:reshape_forw}
    \STATE Send $\widehat{\h}_i \gets \text{rand-SVD}(\hs_\ii, r)$, the rank $r$ approximation of $\hs_\ii$, to server
    \STATE $\h_\ii\gets$ Reshape $\widehat{\h}_i$ back to $\mathbb{R}^{\xdim \times \M}$ at the server \label{line:randsvd_and_send}
\ENDFOR
\STATE  \textbf{Return} $G\gets \Big({\mfrac{1}{|\B|}\sum_{\ii}\h_\ii}\Big)^\top\Big(\mfrac{1}{|\B|}\sum_{\ii}\h_\ii\Big)$ \label{line:GramJacobEstimate}
\end{algorithmic}
\end{algorithm}

We present the $\approxg$ procedure in Algorithm~\ref{alg:approxg}. Each participating client $i$ first reshapes its stochastic Jacobian $\hh_i$ from a tall matrix ($\xdim \times \M$) to a square matrix ($\sqrt{\xdim\M} \times \sqrt{\xdim\M}$), padding any missing entries with zeros (\Cref{alg:approxg}, line~\ref{line:reshape_forw}). The reshaped matrices are then compressed using the leading $r$ components of randomized-SVD \citep{randsvd} and sent to the server. Without reshaping, the smallest component communicated to the server would be of size $\xdim\times 1$. Reshaping enables us to achieve further compression. We adjust the number of singular components, $r$ in randomized-SVD, so that the upload budget of each client is $\xdim \times 1$, the same as the size of one model. While this compression technique requires additional computation, using randomized-SVD to compute a small number of components keeps the cost manageable. Also, stochastic gradients are reused in the first local iteration of clients, avoiding extra gradient computations due to \Cref{alg:approxg}.
At the server, the Gram matrix of the Jacobian is then approximated by the Gram matrix of the averaged compressed client Jacobians (line~\ref{line:GramJacobEstimate}). 
Further details are provided in \Cref{app_sect:approxg_details}.

$\our$ addresses the two key issues with $\fedmgda$ \citep{FMGDA}, 1) local training drift, and 2) high communication cost, by using the aggregated stochastic gradients $\sum_k w_k^{(t+1)} \sG \fik$ for local iterations at the clients (line \ref{line:our_local_sgd} in \Cref{alg:our}). This reduces the model drift compared to $\fedmgda$, which carries $M$ independent local updates for the $M$ objectives. Experimental results in Figure~\ref{fig:locit_drift} (\Cref{sect:results_insights}) and Figure~\ref{fig:local_improvements} (Appendix~\ref{app_sect:local_progress_gap}) also illustrate the effect of this phenomenon.

\begin{proposition}[Communication cost comparison]
For the server to calculate the aggregation weights, the clients in $\fedmgda$ \citep{FMGDA} send $\M$ separate updates for the $\M$ objective functions, which results in an upload cost of $\bigtheta{\M\xdim}$ for each participating client per round. In contrast, the clients in $\our$ upload a single averaged model update $\D_\ii^\tp{t}$ (line~\ref{line:our_update_calc} in \Cref{alg:our}), reducing the communication cost to $\bigtheta{\xdim}$. Computing the averaging weights in $\getw$ incurs an additional communication cost of $\bigtheta{\xdim}$ (\Cref{alg:approxg}). Therefore, the per-client communication cost in every round of $\our$ is $\bigtheta{\xdim}$, significantly reducing the communication cost relative to $\fedmgda$.
\end{proposition}

\section{THEORETICAL ANALYSIS}
\label{sect:theory}
For non-convex and smooth objective functions, we demonstrate the convergence of $\our$ to a Pareto stationary solution using stochastic gradients. 

\paragraph{Algorithmic Simplifications.}
For theoretical simplicity, we analyze $\our$ with a slightly modified subroutine $\approxg$ (see \Cref{app_sect:changes_in_alg_theory}).
% \vspace{-3mm}
\begin{itemize}[noitemsep, leftmargin=*]
    \item Clients communicate their stochastic Jacobians using an unbiased compression operator $\Q$ (see Assumption~\ref{assum:compress} below), rather than the more practical but biased randomized-SVD.
    \item To ensure that the output $G$ of $\approxg$ is an unbiased estimator of $\G \lvect \p{\x}^\top \G \lvect \p{\x}$, the server selects two additional sets $\B_1, \B_2$ of $n'$ clients, each sampled uniformly at random. The resulting Gram matrix estimate is given by $\Big( \frac{1}{n'} \sum_{\ii \in \B_1} \Q\Big(\hh_\ii\Big) \Big)^\top \Big( \frac{1}{n'} \sum_{\ii \in \B_2} \Q\Big(\hh_\ii\Big) \Big)$.
\end{itemize}
In addition, in $\getw$ subroutine, we use $K=1$. \camera{The full procedure with all simplifications for theoretical analyses can be found in Algorithm~\ref{alg:our_theory_app} in Appendix~\ref{app_sect:changes_in_alg_theory}.}

Even in the centralized setting, the theoretical convergence of MOO algorithms  has only recently been explored \cite{direction_oriented, fernando2022mitigating, three_way_tradeoff}. In our federated setting, additional challenges arise from local training at the clients, where client drift occurs due to heterogeneous data distribution, partial client participation, and communication cost considerations. 

\paragraph{Assumptions.}
We provide the convergence analysis of $\ourg$ under the following assumptions, which are standard in the FL and MOO literature \citep{fedvarp, direction_oriented, FMGDA, fedpaq}, and milder than those used in the existing literature.

\begin{assump}[Smoothness] \label{assum:smoothness}
The loss functions are $\smo$-smooth, i.e. for all clients $\ii \in [\N]$, tasks $\kk \in [\M]$, and $ \x,\x' \in \mathbb{R}^{\xdim}$, $\norm{\G \fik (\x)-\G \fik (\x')} \leq \smo \norm{\x-\x'}$.
\end{assump}
% \vspace{-3mm}
\begin{assump}[Bounded Variance] \label{assum:lochet}
Local stochastic gradients are unbiased and bounded-variance estimators of true local gradient, i.e., for all $\x \in \mathbb{R}^\xdim$, $\ii \in [\N]$, and $\kk \in [\M]$, $\E[{\sG \fik\p{\x}}] = \G \fik \p{\x}$ and \mbox{$ \E||{\sG\fik\p{\x}-\G\fik\p{\x}}||^2\leq\lhetsq.$} 
\end{assump}
% \vspace{-3mm}
\begin{assump}[Bounded Heterogeneity] \label{assum:globhet}
Average distances of local gradients to the global gradient is bounded, i.e., for all $\x \in \mathbb{R}^\xdim$, and $\kk \in [\M]$, 
% $\E[{\sG \fik\p{\x}}] = \G \fik \p{\x}$ and 
\mbox{$\tfrac{1}{\N}\sum_\ii||\G\fik\p{\x}-\tl_\kk\p{\x}||^2\leq\ghetsq.$} 
\end{assump}
% \vspace{-3mm}
\begin{assump}[Bounded Gradients] \label{assum:bounded_grad}
The local objective gradients have bounded norms, i.e, for all $\x \in \mathbb{R}^\xdim$, $\ii \in [\N]$, $\kk \in [\M]$, $||\G\fik\p{\x}||\leq\bg$ for some $\bg>0$. 
\end{assump}
% \vspace{-3mm}
\begin{assump}[Unbiased Compression] \label{assum:compress}
The random compression operator $\Qp{\cdot}$ is unbiased and has a variance bound growing with the norm of the input, i.e, for all $\x \in \mathbb{R}^\xdim$, $\E[\Qp{\x}]=\x$ and \mbox{$\E||\Qp{\x}-\x||^2\leq\q||\x||^2$}, for some $q>0$. This assumption is satisfied by several common quantizers \citep{fedpaq}.
\end{assump}

\begin{theorem}[Convergence of $\ourg$]
\label{thm:FedCMO}
Suppose Assumptions~\ref{assum:smoothness}-\ref{assum:compress} hold, the client learning rate satisfies $\lrl\leq \frac{1}{2\smo\locit}$, and the server selects $|\B^{(t)}| = \n$ clients in every round. Then the iterates of $\ourg$ satisfy,
\begin{align}
    & \frac{1}{T} \sum_{t=0}^{T-1} \E \norm{\sum_\kk \wk^\tp{t} \G \tl_k  \p{\x^{(t)} } }^2 \leq \underbrace{\bigo{\beta\M\cone^2}}_{\textit{MOO Weight Error}} \notag \\
    & + \underbrace{\bigo{\frac{1}{T\lrg \lrl \locit} + \frac{\smo\lrg\lrl\lhetsq}{\n}}}_{\substack{\textit{Centralized Optimization Error} \\ \textit{for Scalarized Loss}}} + \underbrace{\bigo{\smo\lrg\lrl\locit\bg^2}}_{\substack{\textit{Partial Participation} \\ \textit{Error}}} \nn \\
    &+ \underbrace{\bigo{\smo \lrl (\locit \bg^2 + \sqrt{\locit}\bg\lhet)}}_{\textit{Local Drift Error}}, \label{eq:theorem1}
\end{align}
where 
\mbox{$\cone\triangleq \mco \p{\mfrac{\q+1}{\np}\lhetsq +\mfrac{\q\bg^2}{\np}+\mfrac{\N-\np}{\np(N-1)}\ghetsq + \bg^2}$} is a constant independent of $T$ and $\M$.
\end{theorem}

\textbf{Proof.} See  \Cref{app_sect:proof}.

The convergence bound of $\our$ is decomposed according to the different sources of error.
The step-size $\beta$ in the $\getw$ subroutine affects the error due to the changing MOO weights. If $\beta$ is set to $0$, the algorithm reduces to \texttt{FedAvg} \citep{fedavg}, with the scalar loss $\sum_k w_k^{(0)} \tl_k$.
Setting $\beta = 0$ in (\ref{eq:theorem1}) also recovers the same dominant terms found in the FedAvg bound \citep{fedvarp}. However, in practice, dynamically adjusting the objective weights helps avoid gradient conflicts (see \Cref{tab:scalarized_comparison} in \Cref{sec:experiments}).

\begin{cor}[Convergence Rate]
\label{cor:FedCMOO}
With the client and server learning rates \mbox{$\lrl=\frac{1}{\smo\locit\sqrt{\locit T}}$}, \mbox{$\lrg=\sqrt{\locit}$}, and step size of \mbox{$\getw$ $\beta = \frac{1}{\M \sqrt{T}}$}, the bound on $\frac{1}{T}\sum_{t=0}^{T-1} \E \norm{\sum_\kk \wk^\tp{t} \G \tl_k \p{\x^\tp{t} } }^2$ in (\ref{eq:theorem1}) reduces to,
\begin{align}
    \bigo{\frac{\bg^2 + \cone^2}{\sqrt{T}}} + \bigo{\frac{\lhetsq}{\locit\sqrt{T}} \p{\frac{1}{\n} + \frac{1}{\locit}}}. \nn
\end{align}
\end{cor}

$\our$ converges to a Pareto stationary solution at the rate $\mco(1/\sqrt{T})$, which matches the best-known result for single-objective FL \citep{fedvarp} under similar conditions.

The number of communication rounds needed to achieve $\E \| \sum_\kk \wk \G \tl_k \|^2 \leq \epsilon$ is $T=\mco(1/\epsilon^2)$. The corresponding \textit{sample complexity} of $\our$, which measures the total number of stochastic gradient computations required by the algorithm, is $T \locit \cdot \M \n = \mco(\M \n /\epsilon^2)$, assuming $\locit = \mco(1)$. The $\M$ factor follows since the clients' local updates in $\our$ require computing stochastic gradients for all $\M$ objectives. 
The sample complexity of $\our$ has a better dependence on the number of objectives, $\mco(\M)$, \camera{compared to the prior work.}
In contrast, the complexity of centralized MOO \citep[Theorem 3]{direction_oriented} grows as $\mco(\M^2)$, while that of $\fedmgda$ \citep[Theorem 5]{FMGDA} grows as $\mco(\M^4)$.

Finally, we remove the impractical assumption in $\fedmgda$ \citep{FMGDA}, which requires the variance of stochastic gradients to be bounded by $\bigo{\eta \sigma^2}$, where $\eta$ is chosen as $\mco(1/\sqrt{T})$ in \cite[Assumption 4 and Corollary 6]{FMGDA}. \camera{We provide a detailed discussion of this assumption in Appendix~\ref{app_sect:unusual_assump_in_yang_et_al}.}

\section{PREFERENCE-BASED FEDERATED MOO} \label{sect:preference_alg}
$\our$, like most MOO algorithms in the literature, finds an \textit{arbitrary} Pareto solution. However, users often prefer solutions with specific trade-offs among the different objectives. For example, in healthcare diagnostics, it is desirable to achieve similar detection accuracy across various ethnic groups \citep{yang2022auc}. 
These preferences can be specified in terms of the ratios of the objective function values, as illustrated in the experiment in Figure~\ref{fig:preference_main}. \cite{most,lin_pareto_2019,lin_smooth_2024, epo} study preference-based MOO in the centralized setting. To our knowledge, ours is the first work to address this problem in a federated or distributed setting. Given the preference vector $\rv \in \mathbb{R}_+^{\M}$, we solve,
\begin{equation}
    \begin{aligned}
    \min_{\x\in\mathbb{R}^\xdim} \lvect\p{\x} &:= \sqbr{\tl_1(\x),\tl_2(\x),\dots,\tl_\M(\x)}^\top \\
    \text{subject to} & \quad r_1\tl_1\p{\x} = \dots = r_\M\tl_\M\p{\x}.
    \end{aligned}
    \label{eq:preference_problem}
\end{equation}

In other words, we want the normalized scaled losses $\uv(\rv) \triangleq \tfrac{\rv \odot \lvect \p{\x}}{\sum_\kk r_\kk \tl_\kk\p{\x}}$ ($\odot$ denotes element-wise product) to satisfy $\hat{u}_i \approx 1/M$, for all $i \in [M]$.
To solve (\ref{eq:preference_problem}), inspired by \cite{epo}, we minimize the KL divergence of $\uv(\rv)$ with respect to the uniform distribution, $\mu_\rv = \KL{\uv(\rv)}{\mathbf{1}/M}
% = \sum_k \hat{u}_k \log(M\hat{u}_k)
$. It is shown in \cite{epo} that using the descent direction $\G\lvect\p{\x} \aw$, where \mbox{$a_\kk = r_\kk\p{\log\p{\hat{u}_\kk \M} - \mu_\rv}$}, guarantees a decrease in $\mu_\rv$ at each iteration.
The aggregation weight vector $\w$ solves the problem 
\begin{align}
    \max_{\w\in\WM\cap \W} \w^\top \G\lvect\p{\x}^\top \G\lvect\p{\x}\p{\aw \indic_\mu + \mathbf{1}(1 - \indic_\mu)}, \label{eq:preference_weight_optim}
\end{align}
where $\indic_\mu$ is an indicator function, which takes the value 1 if the KL divergence $\mu$ is greater than a certain threshold. $\WM$ is the probability simplex, while $\W$ is the set of additional practical constraints (see \Cref{app_sect:preference_alg} for more details). The solution to (\ref{eq:preference_weight_optim})
either reduces the KL divergence $\mu_\rv$ or maximizes the total decrease across all objective values. 

Our proposed preference-based method $\ourp$ (see \Cref{app_sect:preference_alg}) is similar to $\our$, with the difference that the aggregation weights are computed using $\getwpref$ (\Cref{alg:getw_G_pref}) instead of $\getw$. 
The per-client communication cost for one round of $\ourp$, like $\our$, is $\bigtheta{\xdim}$.
\setlength{\algorithmicindent}{.52em}  % indentation level
\begin{algorithm}
\caption{$\getwpref$}
\label{alg:getw_G_pref}
\begin{algorithmic}[1]
\STATE \textbf{Input:} preferences $\rv$, losses $\lvect$, Gram matrix $G$
\STATE Compute normalized losses $\uv(\rv) \triangleq \tfrac{\rv \odot \lvect \p{\x}}{\sum_\kk r_\kk \tl_\kk\p{\x}}$
\STATE Compute non-uniformity: $\mu_\rv = \KL{\uv(\rv)}{\frac{\mathbf{1}}{M}}$
\STATE Compute $a_\kk = r_\kk\p{\log\p{\hat{u}_\kk \M} - \mu_\rv}$, $\forall \ \kk \in [\M]$
\STATE \textbf{Return} $\arg\max_{\w\in\WM \cap \W} \w^\top G\p{\aw \indic_\mu + \mathbf{1}(1 - \indic_\mu)}$ 
\end{algorithmic}
\end{algorithm}

\section{EXPERIMENTAL RESULTS}
\label{sec:experiments}

We first describe our experimental setup (\Cref{sect:dataset}), 
followed by our results and insights (\Cref{sect:results_insights}).

\subsection{Datasets and Implementation} \label{sect:dataset}
We consider the following experimental settings:
\begin{enumerate}[noitemsep, leftmargin=*]
\item \textbf{$\mm$}: Using the MNIST dataset \citep{mnist}, we create a two-objective dataset by randomly merging images. The image size is preserved, with two randomly sampled digits (one in the top left and one in the bottom right) in each sample.
\item \textbf{$\mf$}: Similar to the construction of $\mm$, we construct another dataset by combining two randomly sampled images, one from the MNIST dataset and the other from the FashionMNIST dataset \cite{fashionmnist}.
\item \textbf{$\cm$}: We use three-channel images from the CIFAR-10 dataset \citep{cifar10} and place the same randomly selected MNIST digit at a fixed position in all the channels.
\item \textbf{$\ce$} and \textbf{$\cf$}: $\ce$ is a large-scale face dataset, with each sample containing 40 binary attributes, resulting in a problem with 40 objectives. In addition, we construct the $\cf$ dataset with 5 objectives, by partitioning the attributes of $\ce$ into groups of 8. 
\camera{\item \textbf{$\qm$}: The QM9 dataset \citep{qm9dataset} consists of graph representations of molecules, each associated with 11 continuous-valued molecular properties for regression.}
\end{enumerate}

We use CNN-type models with varying sizes for all experiments\camera{, except for $\qm$, where a graph neural network is used}. The smallest model has $34.6$k parameters, while the largest model, ResNet-18, has $11.2$M parameters. All architectures consist of a shared encoder, used by all the objective functions, and objective-specific decoder components, which form the final one or two linear layers and are applied separately for each objective. See \Cref{app_sect:exp_details} for more details.

Each experiment\camera{, except $\qm$,} has $\N=100$ clients, with $10$ selected in each round. 
Since the MNIST, FashionMNIST, and CIFAR-10 datasets have 10 classes each, each sample in the constructed datasets $\mm$, $\mf$, and $\cm$ belongs to one of the $100$ \textit{composite} classes. 
For the Celeb-A datasets, we select the three most balanced attributes
in terms of the $0$/$1$ ratio and generate $8$-class labels using these three binary labels. We follow the Dirichlet distribution with $\alpha=0.3$, as in \cite{feddyn}, to partition the data samples across clients. \camera{For $\qm$ experiments, we use a total of 20 clients, with 4 active in each round. We split the dataset uniformly at random across the clients.} We run all experiments with multiple random seeds on NVIDIA H100 GPUs and present the average results.

We compare the performance of the discussed methods in extensive experiments. We compare $\ouri$ (\Cref{alg:our}), the preference-based $\ourp$ (\Cref{sect:preference_alg}), and prior work $\fedmgda$ \citep{FMGDA}. 
In some experiments, we also examine $\weighted$, which minimizes the average of individual objective losses \citep{direction_oriented, smooth_tchebycheff}.

\subsection{Results and Insights} \label{sect:results_insights}

\begin{figure}[H]
    \centering
    \includegraphics[width=0.49\textwidth]{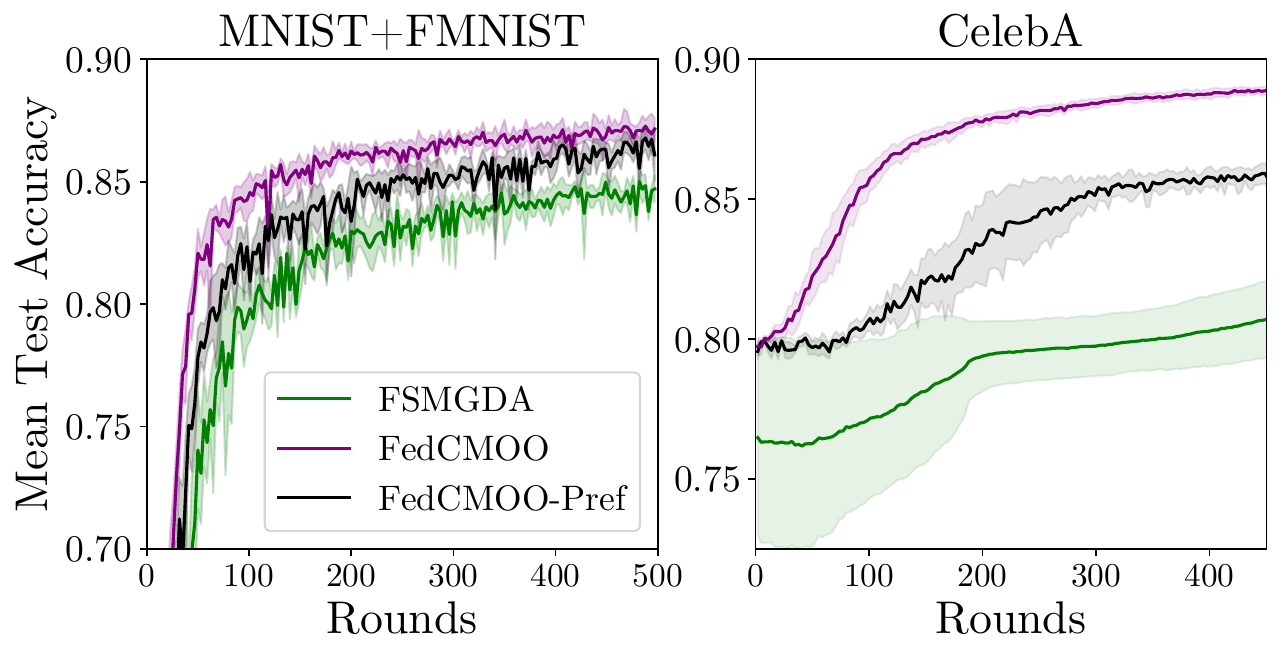}
    \caption{Mean test accuracy with $\mf$ and $\ce$ datasets. $\ouri$ outperforms $\fedmgda$ both in training speed and final accuracy.
    }
    \label{fig:all_lr_test_mean}
\end{figure}

\paragraph{Test Accuracy Curves.} 
We plot the mean test accuracy of the objectives trained together over global rounds in Figure~\ref{fig:all_lr_test_mean}. For $\ourp$, we assign equal importance to all objectives. We observe that our proposed $\our$ achieves faster training. Since $\fedmgda$ performs local training separately for each objective and fuses the gradient information of all objectives only at the end of each round, its performance lags behind due to local model drift. This drift is more pronounced as the number of objectives increases; for example, $\fedmgda$ performs much worse when training on the $\ce$ dataset with 40 objectives. We also provide the final test accuracy for each objective in the two-objective experiments in Table~\ref{tab:loss_L}. As the gap between the difficulty levels of the two objectives increases, $\ourp$ performs better on the more challenging objective. For instance, $\ourp$ performs best on the CIFAR-10 task of $\cm$. However, $\ourp$ requires a higher number of global rounds to train compared to $\our$, as it strives to keep the objective values balanced.
\camera{Please refer to Appendix~\ref{app_sect:training_curves_remaining} for additional training curves from all experiments, Appendix~\ref{app_sect:radial_final_loss_celebs} for radial plots of individual tasks’ final test losses in $\ce$ and $\cf$, and Appendix~\ref{app_sect:delta_m} for $\Delta_M$ results, a commonly used metric in MOO literature \citep{fernando2022mitigating} that measures the average percentage performance loss due to MOO.}

\begin{table}[H]
\caption{The final test accuracy ($\%$) of the first/second objectives in 2-objective settings. The bold values indicate the best accuracy for each objective.}
\label{tab:loss_L}
\begin{tabular}{c|c|c|c}
Experiment                       & \begin{tabular}[c]{@{}c@{}}MNIST+\\ FMNIST\end{tabular} & \begin{tabular}[c]{@{}c@{}}Multi\\ MNIST\end{tabular} & \begin{tabular}[c]{@{}c@{}}CIFAR10+\\ MNIST\end{tabular} \\ \hline
$\fedmgda$ & $93.0/75.4$ & $92.3/88.2$ & $58.2/96.2$ \\
$\ouri$ & $\mathbf{95.5}/78.8$ & $\mathbf{94.4}/\mathbf{92.6}$ & $57.4/\mathbf{96.8}$ \\
$\ourp$ & $94.0/\mathbf{79.2}$ & $93.8/91.1$ & $\mathbf{63.6}/89.1$
\end{tabular}
\end{table}

\begin{figure}[H]
% \vspace{.3in}
\centerline{\includegraphics[width=0.5\textwidth]{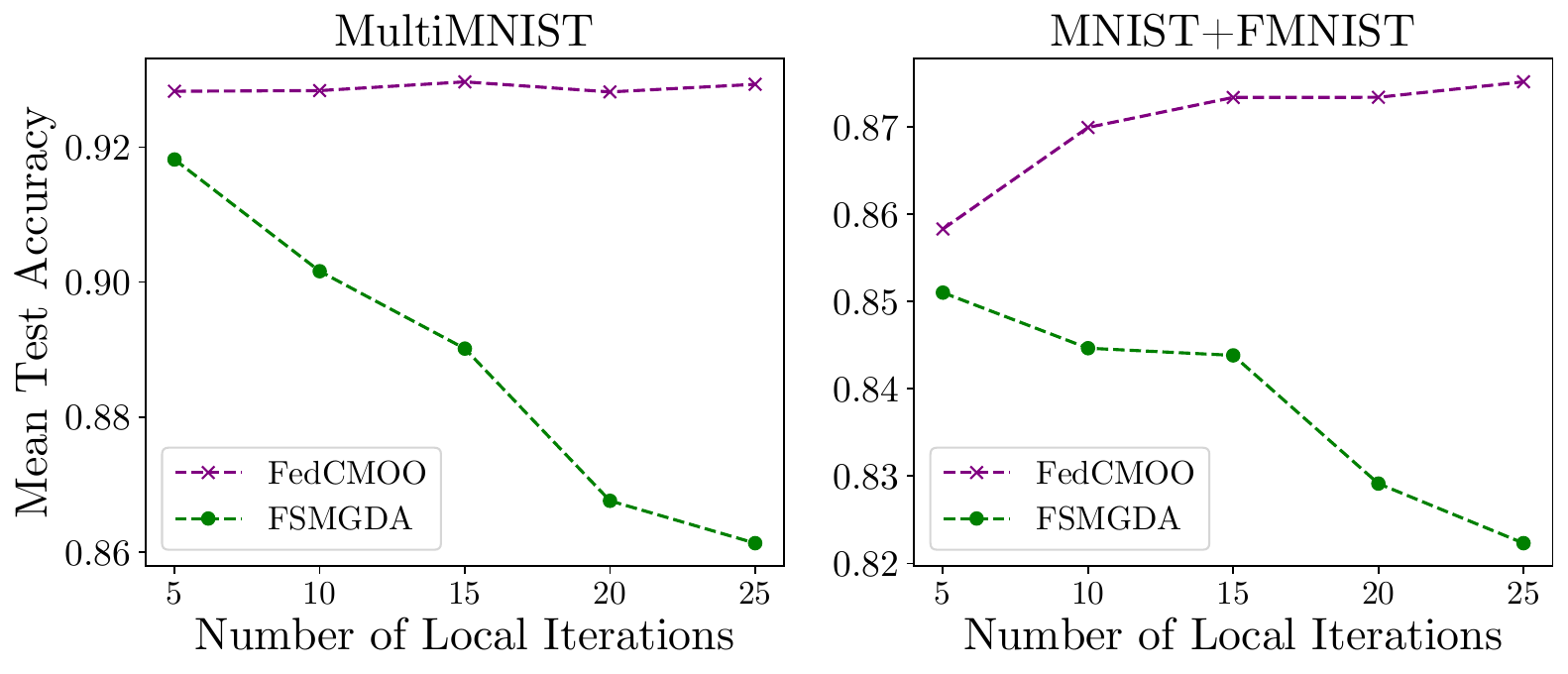}}
    \caption{Mean test accuracy after 500 global rounds. As discussed earlier, with an increasing number of local iterations, the local objective drift of $\fedmgda$ leads to worse performance compared to $\our$.}
    \label{fig:locit_drift}
\end{figure}

\paragraph{Local drift of $\fedmgdanormal$.}
We design an experiment to further validate our claim that $\fedmgda$ suffers from local objective drift. In \Cref{fig:locit_drift}, we compare the mean test accuracy levels achieved by $\our$ and $\fedmgda$ on the $\mm$ and $\mf$ datasets with varying numbers of local training iterations. We observe that the drift caused by the separate training of each local objective in $\fedmgda$ results in worse accuracy. We also compare the per-round local training progress made by $\our$ and $\fedmgda$ (see the result in \Cref{app_sect:local_progress_gap}).
The drift across objectives in $\fedmgda$ hinders the performance of local training.

\begin{figure}[H]
\centerline{\includegraphics[width=0.49\textwidth]{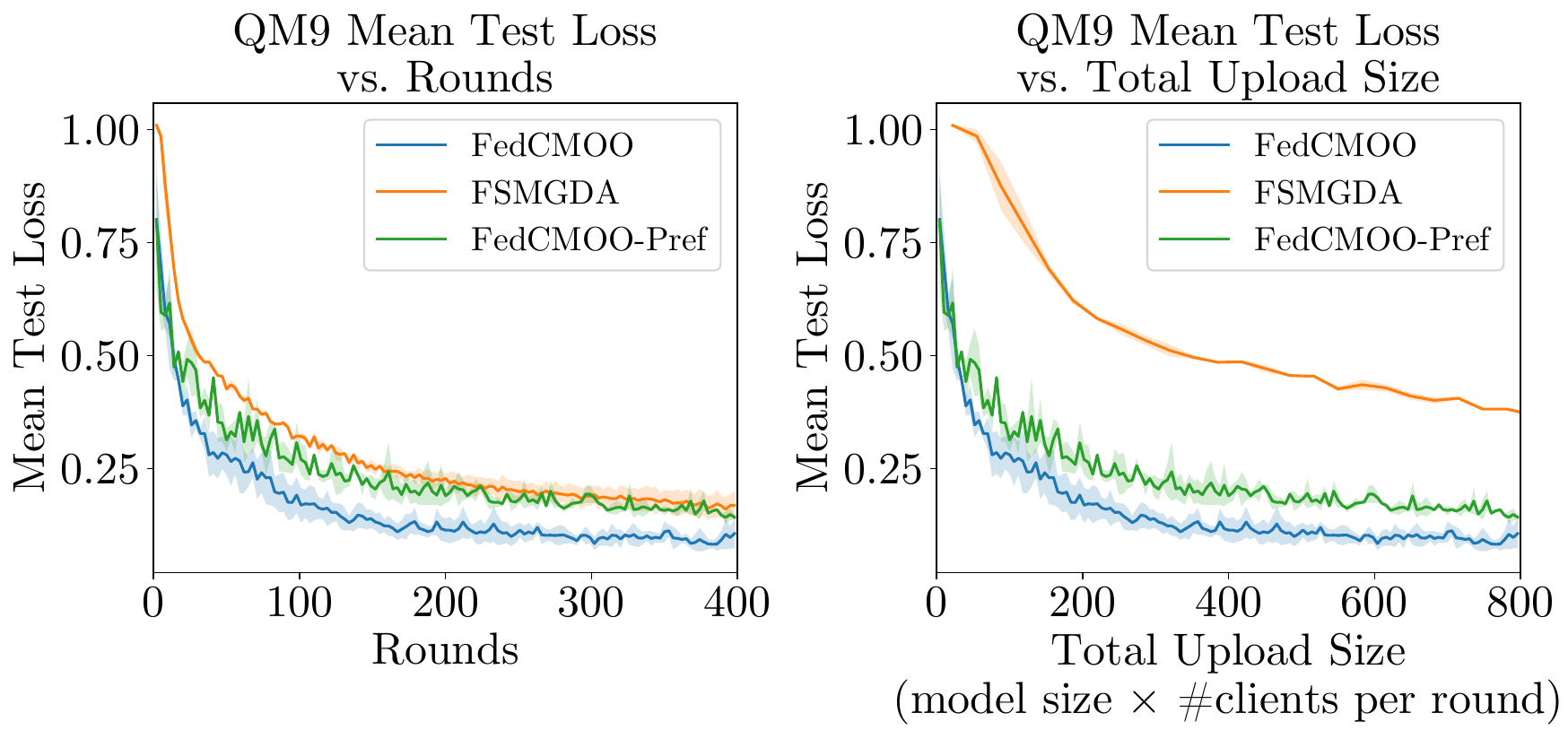}}
    \caption{\camera{Mean test loss curves of $\our$, $\ourp$ (uniform preference), and $\fedmgda$ across 11 $\qm$ tasks. Left: Test loss vs. rounds. Right: Test loss vs. uploaded data (\textit{model size} $\times$ \textit{number of clients per round}). $\our$ outperforms $\fedmgda$ due to reduced local drift, while both $\our$ and $\ourp$ achieve superior communication efficiency.}}
\label{fig:qm9_curves}
\end{figure}

\paragraph{\camera{The communication efficiency of proposed methods.}}
\camera{We present the average test loss curves across 11 regression tasks of the $\qm$ dataset in \Cref{fig:qm9_curves}. The results confirm the superiority of $\our$ over the baseline $\fedmgda$ in terms of training speed relative to the number of server rounds. To further assess the communication efficiency of the proposed methods, we also plot the same results with respect to the amount of uploaded data. The right plot in \Cref{fig:qm9_curves} highlights the significant advantage of our methods, as their communication cost does not scale with the number of objectives, unlike the baseline. Moreover, we present a radial plot of the final test accuracies for each method in \Cref{fig:qm9_radial} in Appendix~\ref{app_sect:qm9_exp_res}, further demonstrating the superiority of $\our$ across individual tasks.} 

\begin{figure}[H]
% \vspace{.3in}
\centerline{\includegraphics[width=0.5\textwidth]{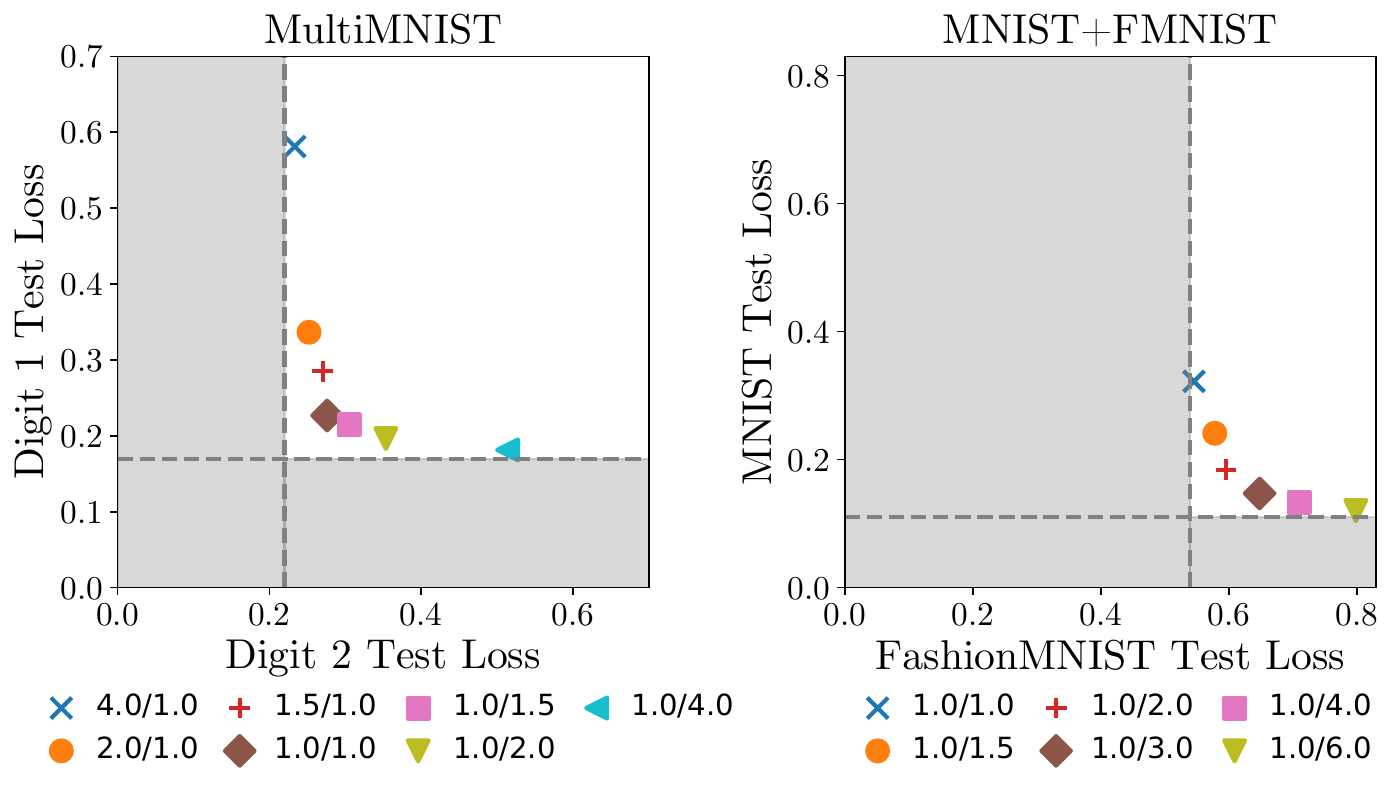}}
    \caption{$\ourp$ effectively finds solutions that align well with user-preferences in most cases. 
    Imbalanced preferences or differences in the difficulty levels of objectives may cause misalignment.}
    \label{fig:preference_main}
\end{figure}

\paragraph{Finding Pareto solutions with user preferences.} 
We test the ability of our proposed $\ourp$ to find Pareto solutions with user-specified preferences.
In \Cref{fig:preference_main}, we present the final objective values achieved by $\ourp$ with varying preferences across objectives, as indicated in plot legends, \camera{showing $\ourp$'s success in identifying solutions that align with user-defined preferences}. The vertical and horizontal dashed lines indicate the minimum loss values from single-objective training, meaning the gray regions are infeasible. When the difficulty gap between the objectives is smaller, e.g., $\mm$, the loss values of the trained models follow the specified preferences.
However, preferences with significantly different weights for the two objectives are not fully satisfied, and the performance is closer to the gray region. On the other hand, in $\mf$, where the FashionMNIST objective is inherently harder than the MNIST objective, the resulting model 
reflects the preference trend with a smaller value of the easy MNIST objective.
The federated setting
further exacerbates the difficulty of alignment with the preference vector.

\paragraph{The drawback of using scalarized loss.}
We further compare the performance of MOO-specific algorithms, i.e., $\our$ and $\fedmgda$, with the single-objective vanilla FL algorithm, FedAvg \citep{fedavg}. We scalarize all objectives into a single objective by averaging them, and refer to this approach as $\weighted$. Although linear scalarization methods can find a Pareto stationary point, they do not have the practical benefits of descent direction maximizing the minimum descent  \cite{grad_manipulation_beyond}. We observe that its performance is highly influenced by loss types and scales across objectives. In Table~\ref{tab:scalarized_comparison}, we change the Digit 1 objective of $\mm$ and the MNIST objective of $\mf$ to $\ell_2$-norm loss while using negative log-likelihood (NLL) loss for the other objectives. We find that $\weighted$ focuses primarily on training the first objective, resulting in poor performance on the other objective. Although MOO-specific $\fedmgda$ performs better than $\weighted$, as discussed above, it suffers from local drift. $\our$ outperforms both $\weighted$ and $\fedmgda$.

\begin{table}[H]
\centering
\caption{Final test accuracy ($\%$) comparison showing poor performance of $\weighted$ when different loss types ($\ell_2$ and NLL) are used across objectives. MOO-specific algorithms maintain more balanced results. }
\label{tab:scalarized_comparison}
\resizebox{\columnwidth}{!}{%
\begin{tabular}{c|cc|cc}
\multirow{2}{*}{Methods} & \multicolumn{2}{c|}{$\mm$} & \multicolumn{2}{c}{$\cm$} \\ \cline{2-5} 
                         & Digit 1         & Digit 2       & CIFAR-10      & MNIST      \\ \hline
$\weighted$              & $10.5$       & $93.4$      &   $20.7$         &     $97.6$         \\ \hline
$\fedmgda$               & $94.6$       & $17.7$      &  $64.2$          &   $61.0$            \\ \hline \hline
$\our$                   & $95.0$       & $70.3$      &   $66.3$         &    $80.0$         
\end{tabular}%
}
\end{table}

\section{CONCLUDING REMARKS}
In this work, we study the federated multi-objective optimization (FMOO) problem. We propose $\our$, a novel algorithm that achieves significant communication savings compared to existing methods. Our convergence analysis holds under milder assumptions than those made in prior work and shows that the complexity of $\our$ achieves better dependence on the number of objectives than existing methods. We also propose a variant of $\our$ that allows users to enforce preferences towards specific solutions on the Pareto front, to achieve different trade-offs among the objective values. Our experiments show the efficacy of our proposed methods. Potential future directions include exploring the effect of task characteristics on training in FMOO and investigating memory- and communication-efficient variance reduction techniques for FMOO settings.

\nocite{crawshaw2020multi, fifty2021efficiently, konevcny2016federated, langner2020beyond, lin2019pareto, macleod2020self, momma2022multi, standley2020tasks}

\subsubsection*{\camera{Acknowledgements}}
\camera{This work was partially supported by the US National Science Foundation under grants CNS-2106891 and CNS-2409138 to CJW and CCF 2045694, CNS-2112471, CPS-2111751, and ONR N00014-23-1-2149 to GJ.}

\bibliography{References}

%%%%%%%%%%%%%%%%%%%%%%%%%%%%%%%%%%%%%%%%%%%%%%%%%%%%%%%%%%%%
\section*{Checklist}

% % %%% BEGIN INSTRUCTIONS %%%
% The checklist follows the references. For each question, choose your answer from the three possible options: Yes, No, Not Applicable.  You are encouraged to include a justification to your answer, either by referencing the appropriate section of your paper or providing a brief inline description (1-2 sentences). 
% Please do not modify the questions.  Note that the Checklist section does not count towards the page limit. Not including the checklist in the first submission won't result in desk rejection, although in such case we will ask you to upload it during the author response period and include it in camera ready (if accepted).

% \textbf{In your paper, please delete this instructions block and only keep the Checklist section heading above along with the questions/answers below.}
% % %%% END INSTRUCTIONS %%%

 \begin{enumerate}

 \item For all models and algorithms presented, check if you include:
 \begin{enumerate}
   \item A clear description of the mathematical setting, assumptions, algorithm, and/or model. [Yes, please see \Cref{sec:algo} and \Cref{sect:theory}.]
   \item An analysis of the properties and complexity (time, space, sample size) of any algorithm. [Yes, please see \Cref{sect:theory}.]
   \item (Optional) Anonymized source code, with specification of all dependencies, including external libraries. [Yes, please see Supplementary Materials.]
 \end{enumerate}

 \item For any theoretical claim, check if you include:
 \begin{enumerate}
   \item Statements of the full set of assumptions of all theoretical results. [Yes, please see \Cref{sect:theory}.]
   \item Complete proofs of all theoretical results. [Yes, please see \Cref{app_sect:proof}.]
   \item Clear explanations of any assumptions. [Yes, please see \Cref{sect:theory}.]     
 \end{enumerate}

 \item For all figures and tables that present empirical results, check if you include:
 \begin{enumerate}
   \item The code, data, and instructions needed to reproduce the main experimental results (either in the supplemental material or as a URL). [Yes, please see \href{https://github.com/askinb/FedCMOO}{https://github.com/askinb/FedCMOO}.]
   \item All the training details (e.g., data splits, hyperparameters, how they were chosen). [Yes, please see \Cref{app_sect:exp_details}.]
         \item A clear definition of the specific measure or statistics and error bars (e.g., with respect to the random seed after running experiments multiple times). [Yes, please see \Cref{sec:experiments} and \Cref{app_sect:exp_details}.]
         \item A description of the computing infrastructure used. (e.g., type of GPUs, internal cluster, or cloud provider). [Yes, the experiments are run on our internal clusters with NVIDIA H100 Tensor Core GPUs as indicated in \Cref{sec:experiments}.]
 \end{enumerate}

 \item If you are using existing assets (e.g., code, data, models) or curating/releasing new assets, check if you include:
 \begin{enumerate}
   \item Citations of the creator If your work uses existing assets. [Yes, please see \Cref{sec:experiments} and \Cref{app_sect:exp_details}.]
   \item The license information of the assets, if applicable. [Not Applicable]
   \item New assets either in the supplemental material or as a URL, if applicable. [Not Applicable]
   \item Information about consent from data providers/curators. [Not Applicable]
   \item Discussion of sensible content if applicable, e.g., personally identifiable information or offensive content. [Not Applicable]
 \end{enumerate}

 \item If you used crowdsourcing or conducted research with human subjects, check if you include:
 \begin{enumerate}
   \item The full text of instructions given to participants and screenshots. [Not Applicable]
   \item Descriptions of potential participant risks, with links to Institutional Review Board (IRB) approvals if applicable. [Not Applicable]
   \item The estimated hourly wage paid to participants and the total amount spent on participant compensation. [Not Applicable]
 \end{enumerate}

 \end{enumerate}

%%% Appendix Supplementary  %%%
\clearpage
\appendix
\onecolumn
\aistatstitle{Appendix for\\``Federated Communication-Efficient Multi-Objective Optimization''}

\section{MULTI-OBJECTIVE OPTIMIZATION AND MGDA}
\label{app_sect:moo_and_mgda}

The multi-gradient descent algorithm ($\MGDA$), a seminal work in multi-objective optimization (MOO) by \cite{mgda}, proposes an iterative gradient-based method to find a Pareto stationary solution with a convergence guarantee. In every iteration, $\MGDA$ aims to find a descent direction of model $\x$ that maximizes the minimum descent across the tasks by forming the problem as
\begin{align}
\max_{\mathbf{d}\in\mathbb{R}^\xdim}\min_{\kk\in[\M]}\left\{\frac{1}{\lr}\p{\tl_\kk\p{\x}-\tl_\kk\p{\x-\lr\mathbf{d}}}\right\}. \label{appeq:mgda_problem}
\end{align}
Using the first-order Taylor expansion, we can approximate $\frac{1}{\lr}\p{\tl_\kk\p{\x}-\tl_\kk\p{\x-\lr\mathbf{d}}}\approx\mathbf{d}^\top\G\tl_\kk\p{\x}$. Also, by regularizing the norm square of $\mathbf{d}$, the problem is to compute
\begin{align}
\max_{\mathbf{d}\in\mathbb{R}^\xdim}\min_{\kk\in[\M]}\left\{\mathbf{d}^\top\G\tl_\kk\p{\x}-\frac{1}{2}\norm{\mathbf{d}}^2\right\}. \label{appeq:mgda_problem1}
\end{align}

Noticing that $\min_{\kk\in[\M]}\mathbf{d}^\top\G\tl_\kk\p{\x}-\frac{1}{2}\norm{\mathbf{d}}^2 = \min_{\w\in\WM}\mathbf{d}^\top\p{\sum_{\kk\in [M]}\wk\G\tl_\kk\p{\x}}-\frac{1}{2}\norm{\mathbf{d}}^2$ where $\WM$ is \mbox{$\M$-probability simplex}, \Cref{appeq:mgda_problem1} is equivalent to
\begin{align}
\max_{\mathbf{d}\in\mathbb{R}^\xdim}\min_{\w\in\WM}\left\{ \mathbf{d}^\top\p{\sum_{\kk\in [M]}\wk\G\tl_\kk\p{\x}}  -\frac{1}{2}\norm{\mathbf{d}}^2\right\}. \label{appeq:mgda_problem2}
\end{align}
Since the problem is concave in $\mathbf{d}$ and affine in $\w$, we can switch the order of $\min$ and $\max$ in \Cref{appeq:mgda_problem2} to obtain
\begin{align}
\min_{\w\in\WM}\max_{\mathbf{d}\in\mathbb{R}^\xdim}\left\{ \mathbf{d}^\top\p{\sum_{\kk\in [M]}\wk\G\tl_\kk\p{\x}}  -\frac{1}{2}\norm{\mathbf{d}}^2\right\}. \nn
\end{align}
Now, the solution to $\max_{\mathbf{d}\in\mathbb{R}^\xdim}\left\{ \mathbf{d}^\top\p{\sum_{\kk\in [M]}\wk\G\tl_\kk\p{\x}}  -\frac{1}{2}\norm{\mathbf{d}}^2\right\}$ can be easily found as \mbox{$\mathbf{d}^{*\max} = \sum_{\kk\in [M]}\wk\G\tl_\kk\p{\x}$}. Inserting this into the problem, the problem reduces to find $\w$. We obtain $\MGDA$ problem in \Cref{eq:mgda_obj} in the main text:
\begin{align}
    \w^*= \argmin_{\w \in \WM} \norm{\sum_{\kk\in[\M]} \wk\nabla\tl_\kk\p{\x}}, \label{appeq:mgda_problem3}
\end{align}

where $\WM$ is $\M$-probability simplex. In every iteration of $\MGDA$ \citep{mgda}, the gradients of all objective functions are calculated and then used to find aggregation weights. Then the descent direction is determined as the weighted average of gradients. $\MGDA$ is depicted in \Cref{alg:mgda_app}.

\setlength{\algorithmicindent}{1em}
\begin{algorithm}[H]
\caption{The multi-gradient descent  algorithm ($\MGDA$) by \cite{mgda}}
\label{alg:mgda_app}
\begin{algorithmic}[1]
\STATE \textbf{Input:} Learning rate $\eta$
\STATE \textbf{Initialization:} $\x^\tp{0}\in\mathbb{R}^{\xdim}$
\FOR{$t = 0, 1, \dots, T - 1$}
\STATE Calculate gradients for all objectives, $\G\tl_1\p{\x^\tp{t}}$, $\G\tl_2\p{\x^\tp{t}}$, \dots, $\G\tl_\M\p{\x^\tp{t}}$
\STATE $\w^\tp{t+1}=\argmin_{\w \in \WM} \norm{\sum_{\kk\in[M]} \wk\nabla\tl_\kk\p{\x^\tp{t}}}$
\STATE $\x^\tp{t+1} = \x^\tp{t}-\lr\sum_{\kk\in[M]} \wk^\tp{t+1}\nabla\tl_\kk\p{\x^\tp{t}}$
\ENDFOR
\STATE \textbf{Return} $\x^\tp{T}$ 
\end{algorithmic}
\end{algorithm}

Calculating exact gradients is often not feasible for large-scale applications. Therefore, in practice, stochastic gradients are used in place of exact gradients in \Cref{alg:mgda_app} \citep{stochastic_mgda}. However, introducing stochasticity presents additional theoretical challenges, as using the same random gradients for both weight calculation (\Cref{appeq:mgda_problem3}) and descent introduces bias. \cite{fernando2022mitigating} demonstrate the non-convergence of $\MGDA$ when using stochastic gradients. Recent works have focused on developing provably convergent stochastic MOO algorithms \citep{three_way_tradeoff, fernando2022mitigating, direction_oriented}.

\section{\texorpdfstring{$\approxgnormal$}{ApproxGramJacobian} SUBROUTINE}
\label{app_sect:approxg_details}
The $\approxg$ algorithm is used to approximate the Gram matrix of the Jacobian of the objective loss vector, $\G\lvect\p{\x}^\top\G\lvect\p{\x}$, in a communication-efficient manner in our proposed $\our$ method. Since only stochastic gradients are computed by the participating clients in each round in the FL setting (line~\ref{line:hi_calculation} in Algorithm~\ref{alg:approxg}), the best approximation would involve transmitting those stochastic gradients and calculating the stochastic Gram matrix on the server. However, this would require $\M \times \xdim$ communication, where $\M$ is the number of objectives and $\xdim$ is the size of the model. With $\approxg$, we reduce communication cost while approximating the Gram matrix. We define the stochastic Jacobian of Client $i$ as \mbox{$\hh_i \triangleq \sqbr{\sG\f_{\ii,1}\p{\x},\dots,\sG\f_{\ii,\M}\p{\x}} \in \mathbb{R}^{\xdim \times \M}$}. The goal is to approximate $\frac{1}{(|\B|)^2} \p{\sum_{\ii\in\B}\hh_\ii}^\top \p{\sum_{\ii\in\B}\hh_\ii}$ on the server.

We introduce a compression operator $\Q$, which compresses $\hh_\ii$ into $\h_\ii$ for transmission at a lower communication cost. In our proposed algorithm, we use a randomized-SVD-based compressor, applied after reshaping $\hh_\ii$ into a square matrix by padding it with zeros to handle non-square cases. After the leading components are communicated, the matrix is reconstructed and returned to its original shape. The number of leading components to communicate is selected such that the upload cost is fixed at $\xdim$. We provide two options for $\approxg$:
\begin{enumerate}
    \item \textit{One-way communication}: The first option requires each client $i$ to upload $Q(\hh_i)$, which has an upload size of $\xdim$. After receiving these compressed matrices, the server reconstructs the approximate Jacobian matrix, $\frac{1}{|\B|}\sum_{\ii\in\B} \Qp{\hh_\ii}$, and estimates the Gram matrix as:
    \[G = \p{\frac{1}{|\B|}\sum_{\ii\in\B}\Qp{\hh_\ii}}^\top\p{\frac{1}{|\B|}\sum_{\ii\in\B}\Qp{\hh_\ii}}.\]
    
    \item \textit{Two-way communication}: In most wireless networks, downloads are typically less costly than uploads \citep{fedavg, mcmahan2017federated_blog}. This option requires each client $i$ to perform an additional download of size $\xdim$ to further reduce the approximation error. To achieve this, we first conduct a deeper analysis of the error components in the approximation. Define the compression error of $\hh_\ii$ as $R_\ii$, such that $\hh_\ii = \Q\p{\hh_\ii} + R_\ii = \h_\ii + R_\ii$. \camera{We expand the target Gram matrix below, annotating each term to indicate the communication cost required to obtain the term on the server side.} For the communication analysis, we ignore terms of size $\M$ or $\M^2$ relative to $\xdim$, since $\M^2 \ll \xdim$. For instance, in the $\cf$ experiment, $\M = 5$ and $\xdim = 11.2$M. 

\allowdisplaybreaks{\begin{align}
\p{\sum_{\ii\in\B}\hh_\ii}^\top\p{\sum_{\ii\in\B}\hh_\ii} &= \underbrace{\sum_{\ii\in\B}\hh_\ii^\top\hh_\ii}_{\substack{\M\times\M\\\text{upload}}} + \sum_{\ii\in\B}\sum_{\substack{j\in\B\\j\neq\ii}} \hh_\ii^\top\hh_j \notag \\
& = \sum_{\ii\in\B}{\hh_\ii^\top\hh_\ii} + \sum_{\ii\in\B}\sum_{\substack{j\in\B\\j\neq\ii}} \p{\h_\ii+R_\ii}^\top\p{\h_j+R_j} \notag \\ 
& = \underbrace{\sum_{\ii\in\B}{\hh_\ii^\top\hh_\ii}}_{\substack{\M\times\M\\\text{upload}}} + \underbrace{\sum_{\ii\in\B}\sum_{\substack{j\in\B\\j\neq\ii}} \h_\ii^\top \h_j}_{\substack{\approx\xdim\\\text{upload}}}+ 2\sum_{\ii\in\B}\sum_{\substack{j\in\B\\j\neq\ii}} \h_\ii^\top R_j+ \sum_{\ii\in\B}\sum_{\substack{j\in\B\\j\neq\ii}} R_\ii^\top R_j. \label{math:half_derivation_approxg}
\end{align}}
    
With an upload size of $\M \times \M + \xdim$ per client, the server can exactly recover the first two terms of \Cref{math:half_derivation_approxg}. Additionally, we observe that with an extra upload of size $\xdim$, the third term can also be estimated. To achieve this, the server compresses the sum of the received $\h_j$ matrices, $\sum_{j\in\B} \h_j$, using $\Q$, and sends $\Qp{\sum_{j\in\B} \h_j}$ back to the clients. Each client $\ii$ then uses $\Qp{\sum_{j\in\B} \h_j}$, along with its own $\h_\ii$ and $R_\ii$, to approximately recover the third term in \Cref{math:half_derivation_approxg}, requiring an additional upload of size $\M \times \M$ and a download of size $\xdim$.
\allowdisplaybreaks{\begin{align}
& \Cref{math:half_derivation_approxg} = {\sum_{\ii\in\B}{\hh_\ii^\top\hh_\ii}} + {\sum_{\ii\in\B}\sum_{\substack{j\in\B\\j\neq\ii}} \h_\ii^\top \h_j}+ 2\sum_{\ii\in\B}\sum_{\substack{j\in\B\\j\neq\ii}} \h_\ii^\top R_j+ \sum_{\ii\in\B}\sum_{\substack{j\in\B\\j\neq\ii}} R_\ii^\top R_j \notag \\
&= {\sum_{\ii\in\B}{\hh_\ii^\top\hh_\ii}} + {\sum_{\ii\in\B}\sum_{\substack{j\in\B\\j\neq\ii}} \h_\ii^\top \h_j}+ 2\sum_{\ii\in\B}R_i^\top\sum_{\substack{j\in\B\\j\neq\ii}} \h_j + \sum_{\ii\in\B}\sum_{\substack{j\in\B\\j\neq\ii}} R_\ii^\top R_j \notag \\
&= {\sum_{\ii\in\B}{\hh_\ii^\top\hh_\ii}} + {\sum_{\ii\in\B}\sum_{\substack{j\in\B\\j\neq\ii}} \h_\ii^\top \h_j}+ 2\sum_{\ii\in\B}R_i^\top\p{\sum_{{j\in\B}} \h_j - \h_i} + \sum_{\ii\in\B}\sum_{\substack{j\in\B\\j\neq\ii}} R_\ii^\top R_j \notag \\
& \approx \underbrace{\sum_{\ii\in\B}{\hh_\ii^\top\hh_\ii}}_{\substack{\M\times\M\\\text{upload}}} + \underbrace{\sum_{\ii\in\B}\sum_{\substack{j\in\B\\j\neq\ii}} \h_\ii^\top \h_j}_{\substack{\approx\xdim\\\text{upload}}}+ 2\underbrace{\sum_{\ii\in\B}\underbrace{R_i^\top}_{\substack{\text{already}\\\text{on client }\ii}}\p{\underbrace{\Qp{\sum_{\substack{j\in\B}} \h_j}}_{\substack{\approx\xdim\\\text{download}}} - \underbrace{\h_i}_{\substack{\text{already}\\\text{on client }\ii}}}}_{\substack{\approx \xdim\text{ download}\\\text{+}\M\times\M\text{ upload}}} + \sum_{\ii\in\B}\sum_{\substack{j\in\B\\j\neq\ii}} R_\ii^\top R_j \label{math:final_of_approxg}
\end{align}}

We treat the last term as an error. This two-way communication option requires an upload of size $\p{2\M \times \M + \xdim}$ and a download of size $\xdim$. Since $\M$, the number of objectives, is much smaller than the model dimension, the communication cost is approximately an upload of size $\xdim$ and a download of size $\xdim$ for each participating client per round.
\end{enumerate}

We use randomized-SVD-based compression as $Q$ in our proposed methods. In the main experiments, we utilize the second option with two-way communication. However, we note that the difference in performance, in terms of accuracy and loss, between the two options is minimal. The detailed steps of $\approxg$ are outlined in \Cref{alg:app_approxg}.

\setlength{\algorithmicindent}{1em}
\begin{algorithm}[H]
\caption{$\approxg$ \camera{full procedure}}
\label{alg:app_approxg}
\begin{algorithmic}[1]
\STATE \textbf{Input:} participating client set $\B$, model $\x$, option
\FOR{each client $\ii \in \B$ in parallel}
    \STATE $\hh_\ii \triangleq \sqbr{\sG\f_{\ii,1}\p{\x},\dots,\sG\f_{\ii,\M}\p{\x}} \in \mathbb{R}^{\xdim \times \M}$
    \STATE $\hs_\ii \gets \text{Reshape}(\hh_\ii)$ to $\mathbb{R}^{\sqrt{\xdim\M} \times \sqrt{\xdim\M}}$ 
    \STATE Send $\widehat{\h}_i \gets \text{randomized-SVD}(\hs_\ii, r)$, the rank $r$ approximation of $\hs_\ii$, to server \hfill \# $\xdim$ upload
    \STATE $\h_\ii\gets$ Reshape $\widehat{\h}_i$ back to $\mathbb{R}^{\xdim \times \M}$ at the server 
\ENDFOR
\IF{the first option, \textit{one-way communication}, is used}
\STATE  $G\gets \Big({\mfrac{1}{|\B|}\sum_{\ii}\h_\ii}\Big)^\top\Big(\mfrac{1}{|\B|}\sum_{\ii}\h_\ii\Big)$ at the server
\ELSIF{the second option, \textit{two-way communication}, is used}
\STATE At the server: $\bar{h} \gets \text{Reshape}(\sum_{\ii\in\B}\h_\ii)$ to $\mathbb{R}^{\sqrt{\xdim\M} \times \sqrt{\xdim\M}}$ 
\STATE Server sends $\widehat{h}\gets \text{randomized-SVD}\p{\bar{h}, r}$, the rank $r$ approximation of $\bar{h}$, to clients in $\B$ \hfill \# $\xdim$ download
\FOR{each client $\ii \in \B$ in parallel}
    \STATE $h \gets $ Reshape $\widehat{h}$ back to $\mathbb{R}^{\xdim \times \M}$ 
    \STATE Send $\hh_\ii^\top\hh_\ii$ to the server \hfill \# $\M\times\M$ upload
    \STATE Send $R_\ii^\top\p{h-\h_\ii}$ to server where $R_\ii=\hh_i-\h_\ii$ \hfill \# $\M\times\M$ upload
\ENDFOR
\STATE  $G\gets \mfrac{1}{|\B|^2}\p{\sum_{\ii\in\B}{\hh_\ii^\top\hh_\ii} + \sum_{\ii\in\B}\sum_{\substack{j\in\B\\j\neq\ii}} \h_\ii^\top \h_j + 2\sum_{\ii\in\B}R_i^\top\left(h - \h_i\right)}$ at the server
\ENDIF
\STATE \textbf{Return} $G$
\end{algorithmic}
\end{algorithm}

The rank $r$ used in randomized-SVD-based compression is selected so that each client’s upload budget is $d \times 1$, equivalent to the size of one model. Both the \textit{one-way} and \textit{two-way communication} options require an upload cost of $\xdim$, while the \textit{two-way communication} option also incurs an additional $\xdim$ download for each participating client per round. The total communication cost for both options is $\bigtheta{\xdim}$.

Next, we compare the empirical performance of the two communication options for the proposed $\approxg$. Additionally, we evaluate the performance of $\approxg$ against two commonly used compression methods as alternatives to randomized-SVD-based compression.

\paragraph{Comparison with other compression methods.}
We compare the proposed communication-efficient approximation method, $\approxg$, which uses randomized-SVD-based compression \citep{randsvd}, with top-$k$ sparsification and random masking. In top-$k$ sparsification, only the $k$ elements with the largest absolute values are communicated, with $k$ determined by the compression ratio, considering the size of the index map of the communicated entries as well. In random masking, a subset of entries at random positions is sent, with the number of entries determined by the compression ratio. For a fair comparison, we ensure that the communication costs for randomized-SVD-based compression, top-$k$ sparsification, and random masking are the same. The compared methods are $\approxg$ with \textit{one-} and \textit{two-way communication} options using randomized-SVD-based compression, top-$k$ sparsification, and random masking. Additionally, we consider using the sum of the clients' Gram matrices as an estimate of the global Gram matrix $\p{\sum_\ii \hh_\ii^\top \hh_\ii}$ with a proper scale.

We run $\ourp$ (with uniform preference across objectives) on different datasets, sending all stochastic gradients from clients to the server. 
At each round, we evaluate the reconstruction error of the Gram matrix ($G$) by comparing the ground truth, computed using all stochastic gradients, with the approximations obtained from the different compression methods.
We use normalized root mean squared error (nRMSE) as the error metric, defined as \(\text{nRMSE(ground truth, estimate)} = \frac{\norm{\text{ground truth} - \text{estimate}}_2}{\norm{\text{ground truth}}_2}\). The results are presented in \Cref{app_table:G_approx_error}. We observe that the proposed $\approxg$ with randomized-SVD provides better estimation performance for the Gram matrix ($G$) of the stochastic Jacobian compared to other compression methods as the number of objectives and model dimension grow. Although methods have slight performance differences in the experiments with small number of objectives, randomized-SVD outperforms others in $\ce$ experiment with 40 objectives.  

\begin{table}[H]\centering
\caption{Average nRMSE (\%) of the estimation of Gram matrix of the stochastic Jacobian ($G$) in experiments with \ourp.}\label{app_table:G_approx_error}\begin{tabular}{|c|c|c|c|c|c|} 
 \hline 
Compression Method & $\ce$ & \begin{tabular}{@{}c@{}}CIFAR10+ \\ MNIST\end{tabular} & \begin{tabular}{@{}c@{}}MNIST+ \\ F.MNIST\end{tabular} & $\mm$ & Mean \\ \hline  Sum of client Gram matrices & $50.95$ & $11.51$ & $20.61$ & $14.56$ & $24.41$ \\ \hline
\begin{tabular}{@{}c@{}}$\approxg$ \textit{two-way comm.}  \\ with randomized-SVD\end{tabular} & $10.15$ & $1.38$ & $2.04$ & $1.76$ & $3.83$ \\ \hline
\begin{tabular}{@{}c@{}}$\approxg$ \textit{two-way comm.}  \\ with random masking\end{tabular} & $51.25$ & $10.26$ & $17.40$ & $12.98$ & $22.97$ \\ \hline
\begin{tabular}{@{}c@{}}$\approxg$ \textit{two-way comm.}  \\ with top-$k$ sparsification\end{tabular} & $17.15$ & $0.15$ & $0.04$ & $0.15$ & $4.38$ \\ \hline
\begin{tabular}{@{}c@{}}$\approxg$ \textit{one-way comm.}  \\ with randomized-SVD\end{tabular} & $10.87$ & $1.68$ & $1.79$ & $1.96$ & $4.08$ \\ \hline
\begin{tabular}{@{}c@{}}$\approxg$ \textit{one-way comm.}  \\ with random masking\end{tabular} & $76.54$ & $4.23$ & $6.29$ & $5.70$ & $23.19$ \\ \hline
\begin{tabular}{@{}c@{}}$\approxg$ \textit{one-way comm.}  \\ with top-$k$ sparsification\end{tabular} & $24.13$ & $0.04$ & $0.01$ & $0.10$ & $6.07$ \\ \hline
\end{tabular}\end{table}

\section{$\ourpnormal$ ALGORITHM} 
\label{app_sect:preference_alg}
We provide the details of the proposed preference-based federated communication-efficient algorithm, $\ourp$.
\subsection{The Preference Definition and Evaluation Metric}

We consider an experimental setting where the user specifies the preference vector $\rv\in\mathbb{R}_+^\M$, whose entries are preference weights, $r_1$, $r_2$, \dots, $r_\M$. The aim is to find a Pareto stationary solution $\x$ that satisfies \mbox{$r_1\tl_1\p{x}= r_2\tl_2\p{x}=\dots =r_\M\tl_\M\p{x}$}. Given the preference vector $\rv \in \mathbb{R}_+^{\M}$, the problem of interest is formulated as:
\begin{equation}
    \begin{aligned}
    \min_{\x\in\mathbb{R}^\xdim} \lvect\p{\x} &:= \sqbr{\tl_1(\x),\tl_2(\x),\dots,\tl_\M(\x)}^\top \\
    \text{subject to} & \quad r_1\tl_1\p{\x} = \dots = r_\M\tl_\M\p{\x}. \label{app_eq:epo_problem}
    \end{aligned}
\end{equation}

Inspired by \cite{epo}, we follow a KL divergence-based method to solve \Cref{app_eq:epo_problem}. First, we define the normalized scaled losses of model $\x$ for a given preference vector $\rv$:
\[\uv(\rv) \triangleq \frac{\rv \odot \lvect \p{\x}}{\sum_\kk r_\kk \tl_\kk\p{\x}}\]
where $\odot$ denotes element-wise product. If the constraint of \Cref{app_eq:epo_problem} is satisfied, each entry $\hat{u}_\kk \ (\forall \ \kk\in[\M])$ of $\uv(\rv)$ should be $1/\M$. Then, we define the KL divergence, which measures the distance of $\uv$ from the uniform vector with entries of $1/\M$, i.e. non-uniformity,
\[ \mu_\rv = \KL{\uv(\rv)}{\mathbf{1}/M} = \sum_{\kk\in[\M]} \hat{u}_\kk\log\p{\frac{\hat{u}_\kk}{1/\M}}.\]

\subsection{The Descent Direction and Task Weights} \label{app_sect:epo_des_dir}
By Theorem 1 in \cite{epo}, it is guaranteed that when we have exact gradients $\{\G\tl_\kk\p{\x}\}_{\kk\in[\M]}$, an update step $\x\gets\x-\lr\underbrace{\G\lvect\p{\x}\aw}_{\sum_{\kk\in[\M]}a_\kk\G\tl_\kk\p{\x}}$ does not increase $\mu_\rv$ with a sufficiently small step size $\lr$ where $a_\kk = r_\kk\p{\log\p{\frac{\hat{u}_\kk}{1/\M}}-\mu_\rv}$. 

We use an indicator function, $\indic_\mu$, which is $0$ if $\mu_\rv \leq \epsilon$ and $1$ otherwise, where the threshold $\epsilon = 0.01$ is used in the experiments. To find the update direction using exact gradients, the task weights can be determined by either maximizing the total descent across objectives or decreasing the KL divergence, such that:
\begin{align}
    \max_{\w\in\WM} \w^\top \G\lvect\p{\x}^\top \G\lvect\p{\x}\p{\aw \indic_\mu + \mathbf{1}(1 - \indic_\mu)}. \label{app_eq:plain_epo_opt_problem}
\end{align}

However, solving \Cref{app_eq:plain_epo_opt_problem} may not yield task weights that perform well, as decreasing KL divergence $\mu_\rv$ without any constraints may lead to increases in all objectives, which is undesirable. To address this, first, the following sets are defined:
\begin{enumerate}[noitemsep, leftmargin=*]
    \item $J\triangleq\left\{\kk\mid \p{\G\lvect\p{\x}\aw}^\top\G\tl_\kk\p{\x} > 0 \right\}$: The indices of objectives whose gradients are positively aligned with the direction of decreasing KL divergence, $\G\lvect\p{\x}\aw$. The loss of these objectives is expected to decrease.
    
    \item $\bar{J}\triangleq\left\{\kk\mid \p{\G\lvect\p{\x}\aw}^\top \G\tl_\kk\p{\x} \leq 0 \right\}$: The indices of objectives whose gradients are negatively aligned with or perpendicular to the direction of decreasing KL divergence, $\G\lvect\p{\x}\aw$. The loss of these objectives is expected to increase or remain the same.
    
    \item $J^*\triangleq\left\{\kk\mid \kk \in \arg\max_j r_j\tl_j\p{\x} \right\}$: The indices of objectives for which the preference multiplied by the current loss value is the maximum across all objectives. These are the objectives for which we should not increase the loss.
\end{enumerate}
To ensure not increasing the loss of objectives in $J^*$, we require the solution weights $\w^*$ to satisfy:
\[\p{\G\lvect\p{\x}\w^*}^\top\G\tl_\kk\p{\x}\geq0,\quad\quad\forall\kk\in J^*.\]
Additionally, to prevent a potential increase in the loss functions of all objectives when $J$ is an empty set, we introduce the following constraint. We define $\indic_{J\neq\varnothing}$ as an indicator function that returns $0$ if $J$ is empty and $1$ otherwise:
\[ \p{\G\lvect\p{\x}\w^*}^\top\G\tl_\kk\p{\x}\geq \p{\G\lvect\p{\x}\aw}^\top\G\tl_\kk\p{\x}\indic_{J\neq\varnothing} ,\quad\quad\forall\kk\in \bar{J} \ \backslash \ J^*. \]
These two constraints form $\W$, the practical constraints mentioned in Problem~(\ref{eq:preference_weight_optim}) in the main text.  Incorporating these constraints into Problem~(\ref{app_eq:plain_epo_opt_problem}), we obtain:
\begin{align}
    \text{Problem~(\ref{eq:preference_weight_optim})} \equiv \max_{\w\in\WM} & \ \w^\top \G\lvect\p{\x}^\top \G\lvect\p{\x}\p{\aw \indic_\mu + \mathbf{1}(1 - \indic_\mu)} \label{app_eq:epo_full_exact_grad} \\
    \text{s.t.} \quad & \p{\G\lvect\p{\x}\w}^\top\G\tl_\kk\p{\x} \geq 0, \quad \forall\kk\in J^*, \nn \\
    & \p{\G\lvect\p{\x}\w}^\top\G\tl_\kk\p{\x} \geq \p{\G\lvect\p{\x}\aw}^\top\G\tl_\kk\p{\x}\indic_{J\neq\varnothing}, \quad \forall\kk\in \bar{J} \ \backslash \ J^*. \nn
\end{align}
We refer the reader to \cite{epo} for further explanations and insights.

\subsection{Preference-based Federated Communication-Efficient MOO}
Since exact gradients cannot be calculated and communication is limited in our federated problem setting, \Cref{app_eq:epo_full_exact_grad} cannot be solved exactly. Instead, similar to $\our$, we first approximate the Gram matrix of the Jacobian, $G\approx\G\lvect\p{\x}^\top\G\lvect\p{\x}$, on the server. Clients also share their loss values, which consist of only $\M$ scalar values per client, with the server, allowing the server to estimate the global objective loss values by averaging them. With these approximations, all the steps presented in \Cref{app_sect:epo_des_dir} can be performed approximately on the server in a communication-efficient way.

First, $\ourp$ uses the same $\approxg$ subroutine as $\our$ to obtain \mbox{$G = [\gw_1, \gw_2, \dots, \gw_\M] \approx \G\lvect\p{\x}^\top\G\lvect\p{\x}$} on the server, where $\gw_1, \gw_2, \dots, \gw_\M$ are the columns of $G$. Additionally, using the local loss values of the participating clients, the objective value estimates, $\widetilde{\tl}_1\p{\x}$, \dots, $\widetilde{\tl}_\M\p{\x}$, are computed. The server then calculates the normalized losses using these objective value estimates. Subsequently, the KL divergence $\mu_\rv$ and the $\aw$ weights are calculated. Note that the sets $J$, $\bar{J}$, and $J^*$ can be approximated using only the columns of matrix $G$ and the estimated loss values, without needing individual objective gradients, as follows:
\allowdisplaybreaks{\begin{align}
    &J = \left\{\kk\mid \aw^\top \gw_\kk > 0 \right\}, \quad\quad 
    \bar{J} = \left\{\kk\mid \aw^\top \gw_\kk  \leq 0 \right\} \quad\quad 
    J^* = \left\{\kk\mid \kk \in \arg\max_j r_j\widetilde{\tl}_j\p{\x} \right\}  \label{app_eq:calculate_j_sets_approx}
\end{align}}

Then the following approximation to Problem~(\ref{app_eq:epo_full_exact_grad}) is solved on the server to find task weights. 
\begin{align}
    \max_{\w\in\WM} & \ \w^\top G \p{\aw \indic_\mu + \mathbf{1}(1 - \indic_\mu)} \label{app_eq:approximate_epo_solve} \\
    \text{s.t.}  \quad & \w^\top\gw_\kk \geq 0, \quad \forall\kk\in J^*, \nn \\
    & \w^\top \gw_\kk \geq \aw^\top \gw_\kk \indic_{J\neq\varnothing}, \quad \forall\kk\in \bar{J} \ \backslash \ J^*. \nn
\end{align}

We depict the $\ourp$ in \Cref{alg:ourp}, which differs from $\our$ only in how the task weights are obtained. Additionally, we provide a detailed presentation of the $\getwpref$ subroutine in \Cref{alg:getw_G_pref}, with more details available in \Cref{app_alg:getwpref_detailed}.

\begin{algorithm}[H]
\caption{\ourp}
\label{alg:ourp}
\begin{algorithmic}[1]
\STATE \textbf{Input:} client and server learning rates $\lrl, \lrg$, number of local steps $\locit$, preference vector $\rv$
\STATE \textbf{Initialize:} global model $\x^\tp{0} \in \mathbb{R}^\xdim$, and task weights $\w^\tp{0}\gets [1/\M,\dots,1/\M]^\top \in \WM$
\FOR{$t = 0, 1, \dots, T - 1$}
    \STATE Select the client set $\B^\tp{t}$ uniformly at random from $[N]$; send $\x^\tp{t}$ to the clients in $\B^\tp{t}$ 
    \STATE $G^\tp{t} \gets \approxg (\B^\tp{t}, \x^\tp{t})$ at the server, where $G^{(t)} \approx \G \lvect (\x^{(t)})^\top \G \lvect (\x^{(t)})$ 
    \STATE $\widetilde{\lvect}^\tp{t}\gets$ Average loss values of participating clients in $\B^\tp{t}$ for all objectives
    \STATE Compute $\w^\tp{t+1} \gets \getwpref$($\rv$, $\widetilde{\lvect}^\tp{t}$, $G^\tp{t}$) at the server and send to clients in $\B^\tp{t}$ \label{line0:fedcmoo_pref} 
    \FOR{each client $\ii \in \B^\tp{t}$ \textbf{in parallel}}
        \STATE Initialize local model: $\x_\ii^\tp{t,0} \gets \x^\tp{t}$ 
        \FOR{$r = 0, \dots, \locit-1$}
            \STATE 
            \(
            \x_\ii^\tp{t,r+1} \gets \x_\ii^\tp{t,r} - \lrl \sum_{k=1}^\M \wk^\tp{t+1} \sG\fik (\x_\ii^\tp{t,r})
            \) 
        \ENDFOR 
        \STATE Send
        \(
        \D_\ii^\tp{t} \triangleq \mfrac{1}{\locit \lrl} \big( \x^\tp{t} - \x_\ii^\tp{t,\locit} \big)
        \) to the server 
    \ENDFOR
    \STATE
    \(
    \x^\tp{t+1} \gets \x^\tp{t} - \lrg \lrl \locit \mfrac{1}{|\B^\tp{t}|} \sum_{\ii \in \B^\tp{t}} \D_\ii^\tp{t}
    \) 
\ENDFOR
\STATE \textbf{Return} $\x^\tp{T}$ 
\end{algorithmic}
\end{algorithm}

\begin{algorithm}[H]
\caption{$\getwpref$ in detail} \label{app_alg:getwpref_detailed}
\begin{algorithmic}[1]
\STATE \textbf{Input:} preferences $\rv$, losses $\lvect$, Gram matrix $G = [\gw_1, \gw_2, \dots, \gw_\M]$
\STATE Compute normalized losses $\uv(\rv) \triangleq \tfrac{\rv \odot \lvect \p{\x}}{\sum_\kk r_\kk \tl_\kk\p{\x}}$
\STATE Compute non-uniformity: $\mu_\rv = \KL{\uv(\rv)}{\frac{\mathbf{1}}{M}}$
\STATE \camera{Compute $\aw\in\mathbb{R}^\M$ such that  $a_\kk = r_\kk\p{\log\p{\frac{\hat{u}_\kk}{1/\M}}-\mu_\rv}$ for all $\kk\in[\M]$}
\STATE $J \gets \left\{\kk\mid \aw^\top \gw_\kk > 0 \right\}$
\STATE $ \bar{J} \gets \left\{\kk\mid \aw^\top \gw_\kk  \leq 0 \right\}$
\STATE $ J^* \gets \left\{\kk\mid \kk \in \arg\max_j r_j\widetilde{\tl}_j\p{\x} \right\} $
\STATE $\w \gets$ a solution to optimization problem (\ref{app_eq:approximate_epo_solve})
\STATE \textbf{Return} $\w$
\end{algorithmic}
\end{algorithm}

\paragraph{Practical improvement.} We observe that enforcing a minimum weight on every task accelerates training with the $\ourp$ method, even when this minimum weight is small. In the $\ourp$ experiments, except for \Cref{fig:preference_main}, where the goal is to find a Pareto solution aligned with specific preference, we project the output of line~\ref{line0:fedcmoo_pref} in \Cref{alg:ourp}, $\wtp$, onto a vector where each entry is at least $\tfrac{1}{5 \times \M}$, where $\M$ is the number of objectives.

\section{EXPERIMENTAL DETAILS} 
\label{app_sect:exp_details}

We conduct simulated experiments in a federated multi-objective optimization setting, primarily using PyTorch and CVXPY packages on our internal clusters equipped with NVIDIA H100 Tensor Core GPUs. In this section, we describe the experiments in detail.

\subsection{Datasets} \label{app_sect:datasets}
We use \camera{six} different datasets in the experiments:

\begin{enumerate}%[noitemsep, leftmargin=*]
\item \textbf{$\mm$}: We create our $\mm$ dataset using the well-known MNIST dataset \citep{mnist}. The images are maintained at $28\times28$ pixels with 1 channel. For each sample, we place two random digits, one at the top-left and the other at the bottom-right, after cropping and resizing them. Each sample is assigned a new label that combines the labels of the two digits, resulting in 100 unique labels. Samples from this dataset are shown in \Cref{app_fig:mm_samples}.
\begin{figure}[H]
    \centering
    \includegraphics[width=0.65\textwidth]{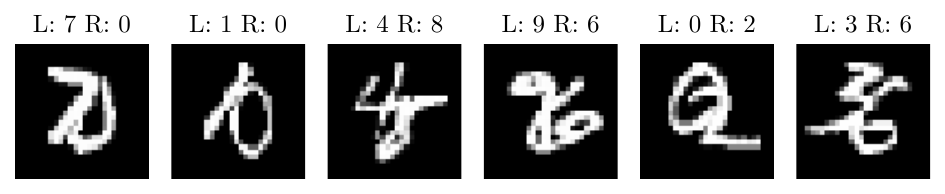}
    \caption{Samples from created $\mm$ dataset. The captions indicate the individual digits of the left (L) and right (R) samples merged.}
    \label{app_fig:mm_samples}
\end{figure}
\item \textbf{$\mf$}: Similar to the construction of $\mm$, we create another dataset by combining two randomly sampled images: one from the MNIST dataset and one from the FashionMNIST dataset \citep{fashionmnist}. The labeling process follows the same approach as in $\mm$. Samples from this dataset are shown in \Cref{app_fig:mf_samples}.
\begin{figure}[H]
    \centering
    \includegraphics[width=0.65\textwidth]{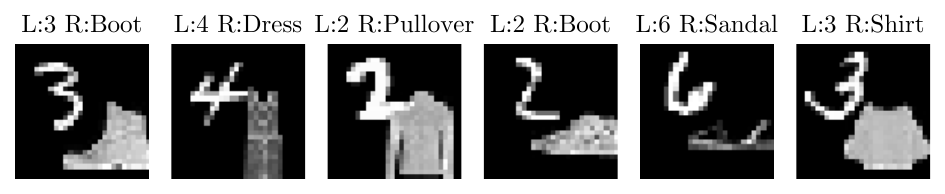}
    \caption{Samples from created $\mf$ dataset. The captions indicate the individual labels of the MNIST (L) and FashionMNIST (R) samples merged.}
    \label{app_fig:mf_samples}
\end{figure}
\item \textbf{$\cm$}: We use three-channel images from the CIFAR-10 dataset \citep{cifar10} and randomly selected MNIST digits. Each digit is cropped, resized, and placed in the center of all channels. The labeling process is similar to that of $\mm$ and $\mf$. Samples from this dataset are shown in \Cref{app_fig:cm_samples}.
\begin{figure}[H]
    \centering
    \includegraphics[width=0.75\textwidth]{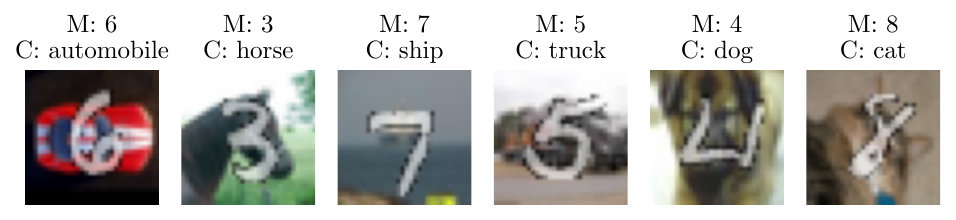}
    \caption{Samples from created $\cm$ dataset. The captions indicate the individual labels of MNIST (M) and CIFAR-10 (C) samples merged.}
    \label{app_fig:cm_samples}
\end{figure}
\item \textbf{$\ce$} and \textbf{$\cf$}: $\ce$ is a large-scale face dataset \citep{celeba}, where each sample contains 40 binary attributes, resulting in a 40-objective problem. We also construct the $\cf$ dataset with 5 objectives by partitioning the 40 attributes of $\ce$ into 5 groups of 8 attributes each. The attributes are grouped based on visual and facial characteristics, ensuring a balanced representation of different facial features in each group. The groups are as follows:\\
Group 1: 
5\_o\_Clock\_Shadow, Blond\_Hair, Bags\_Under\_Eyes, Eyeglasses, Mustache, Wavy\_Hair, Oval\_Face, Rosy\_Cheeks,
\\
Group 2: 
Arched\_Eyebrows, Black\_Hair, Big\_Nose, Blurry, Goatee, Straight\_Hair, Pale\_Skin, Wearing\_Earrings,
\\
Group 3: 
Attractive, Brown\_Hair, Bushy\_Eyebrows, Double\_Chin, Mouth\_Slightly\_Open, Receding\_Hairline, Wearing\_Hat, Young,
\\
Group 4: 
Bald, Big\_Lips, Chubby, Heavy\_Makeup, High\_Cheekbones, Male, Sideburns, Smiling,
\\
Group 5: 
Bangs, Gray\_Hair, Narrow\_Eyes, No\_Beard, Pointy\_Nose, Wearing\_Lipstick, Wearing\_Necklace, Wearing\_Necktie.

In both datasets, we assign each sample with new composite labels using the binary value combinations of attributes `Attractive', `Male', `Mouth\_Slightly\_Open'. In total there are 8 possible new composite labels.

\camera{\item \textbf{\qm}: QM9 is a widely used 11-task regression benchmark \citep{qm9dataset}. We follow the implementations in \cite{lin2023libmtl, fl_molecule}. The QM9 dataset consists of graph data for molecules, with 40k samples used for training and 20k samples for validation and testing, randomly split. The training samples are distributed to the clients uniformly at random. All 11 tasks involve predicting continuous-valued molecular properties. A graph neural network model as described in \cite{lin2023libmtl} is used as the backbone, and linear models are employed as decoders. Mean-squared error is used both as the loss function for training and as the evaluation metric.}

\end{enumerate}

\subsection{Federated Setting Details and Data Partitioning}
Our simulated federated setting consists of $\N = 100$ clients\camera{, except in $\qm$ experiments, where we use $\N = 20$ clients}. To create data heterogeneity across clients in the experiments, we use a Dirichlet distribution over the newly assigned composite labels. We follow this distribution to determine the number of samples each client receives from each label. We follow the implementation from \cite{feddyn}. \camera{In $\qm$ experiments, data samples are distributed uniformly at random across clients.} All clients are assigned an equal number of data points.

\subsection{Models, Training, and Hyperparameter Selection}

We use convolutional neural network (CNN) models of varying sizes in all experiments\camera{, except for $\qm$, where we use a graph neural network (GNN)}. For $\mm$ and $\mf$, we employ a LeNet-like CNN encoder \citep{lenet, senermtl} with two linear layers as the decoders, resulting in 34.6k parameters. In $\cm$, we use a CNN encoder adapted from \cite{feddyn} with a single linear layer as the decoders, totaling 1.73M parameters. For $\ce$ and $\cf$, we utilize a ResNet-based encoder adapted from \cite{fedexp}, paired with one linear layer as the decoders, resulting in 11.2M parameters. \camera{In $\qm$, we use a GNN model as described in \cite{lin2023libmtl} as the backbone, with linear decoders, totaling 617k parameters.}

We use negative log-likelihood loss for $\mm$, $\mf$, and $\cm$, and binary cross-entropy loss for the CelebA experiments unless otherwise specified. For $\mm$, $\mf$, and $\cm$, we apply random rotations as data augmentation during training. The batch size is fixed at 128, and we perform 10 local training steps per round unless otherwise stated. Stochastic Gradient Descent (SGD) without momentum is used as the optimizer. \camera{For $\qm$, mean-squared error is used as the loss function, Adam optimizer is employed, and 14 local training steps are performed per round.}

All datasets \camera{except for $\qm$} have predefined training and test splits, which we use along with the default validation splits, when available. If a validation split is not provided, we randomly split the training set, using 80\% for training and 20\% for validation. \camera{For $\qm$, we use 40k training samples and 20k validation and test samples selected randomly.}

We use constant learning rates (without changing during the training) in experiments. To select the learning rates, we run the methods on each experimental setting and evaluate them on the validation set. 

For the $\ce$ experiment, the global learning rate ($\lrg$) is set to $1$, and the local learning rate ($\lrl$) is $0.2$. In the CelebA5 experiment, the global learning rate is $1.6$, and the local learning rate is $0.3$.

For the $\mm$ and $\mf$ experiments, the learning rates are the same but vary depending on the algorithm: for $\fedmgda$, the global learning rate is $2$, and the local learning rate is $0.1$; for $\our$, the global learning rate is $1.2$, and the local learning rate is $0.5$; for $\ourp$, the global learning rate is $1.6$, and the local learning rate is $0.3$.

For, the $\cm$ experiment, the global learning rate is $1.6$, and the local learning rate is $0.3$ for all methods. \camera{Finally, for the $\qm$ experiment, the global learning rate is set to $1$, and the local learning rate is $0.01$ for all methods.}

% \subsection{BASELINE ALGORITHMS} \label{app_sect:baseline_algos}

% \subsection{HYPERPARAMETER SELECTION} \label{app_sect:hyperparams}

\section{ADDITIONAL EXPERIMENTAL RESULTS} \label{app_sect:extra_experimental_results}
We present the additional experimental results here which are not included in the main text due to space limitation.

\subsection{The Training Curves Of The Experiments} \label{app_sect:training_curves_remaining}
\begin{figure*}[h]
    \centering
    \includegraphics[width=1\textwidth]{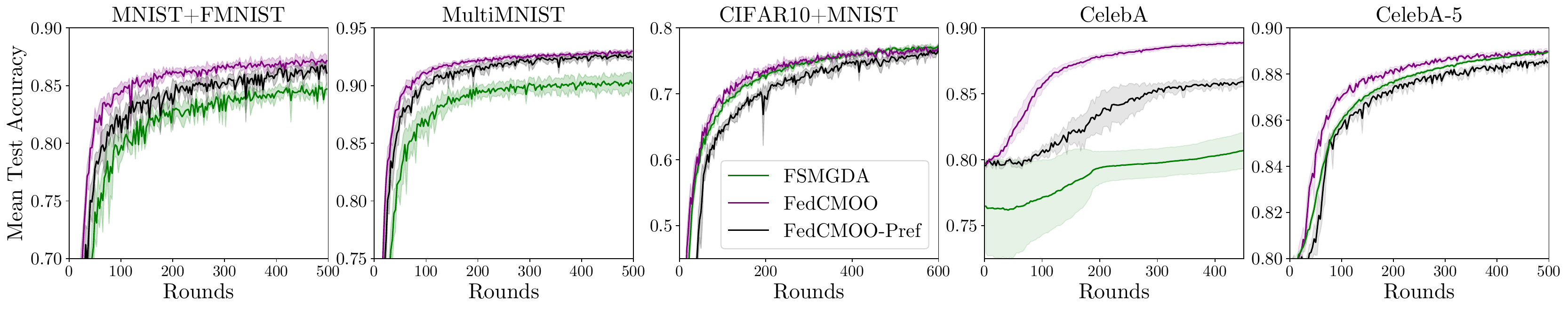}
    \caption{Mean test accuracy in $\mf$, $\mm$, $\cm$, $\ce$, and $\cf$ datasets. $\our$ outperforms the $\fedmgda$ in training speed and final accuracy. $\ourp$ with uniform preference either outperforms or is surpassed by $\fedmgda$ in terms of mean accuracy, but $\ourp$ trains the model for all objectives more uniformly across all tasks (see Table~\ref{tab:loss_L} in the main text).}
    \label{fig:all_lr_test_mean_all}
\end{figure*}

\begin{figure*}[h]
    \centering
    \includegraphics[width=1\textwidth]{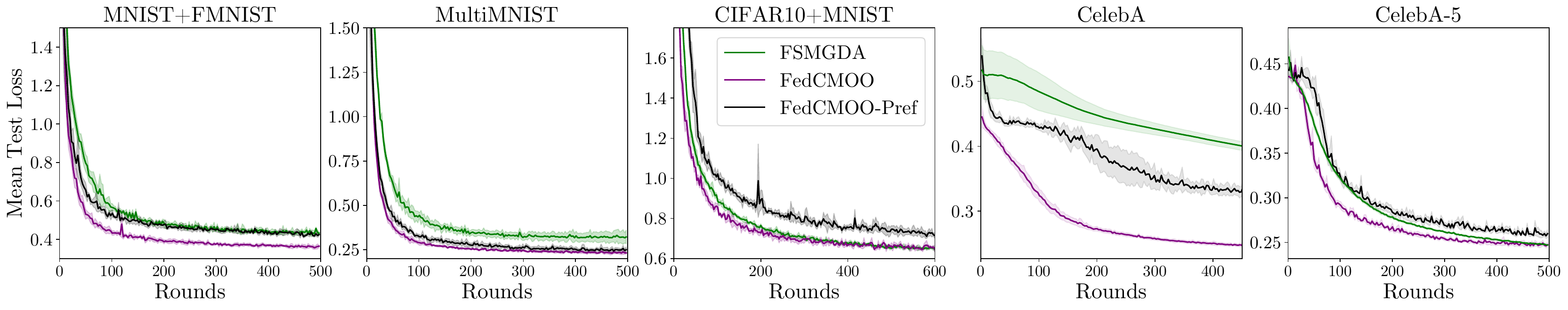}
    \caption{Mean test loss in $\mf$, $\mm$, $\cm$, $\ce$, and $\cf$ datasets. $\ouri$ outperforms the $\fedmgda$ in training speed and final loss.}
    \label{fig:all_lr_loss_test_mean_all}
\end{figure*}

We show the mean test accuracy and mean test loss curves of the compared methods (similar to the experiment in \Cref{fig:all_lr_test_mean} in the main text) in Figures~\ref{fig:all_lr_test_mean_all} and \ref{fig:all_lr_loss_test_mean_all}.

\subsection{The Final Loss Values Of $\ce$ And $\cf$}\label{app_sect:radial_final_loss_celebs}
\begin{figure}[H]
\centerline{\includegraphics[width=0.7\textwidth]{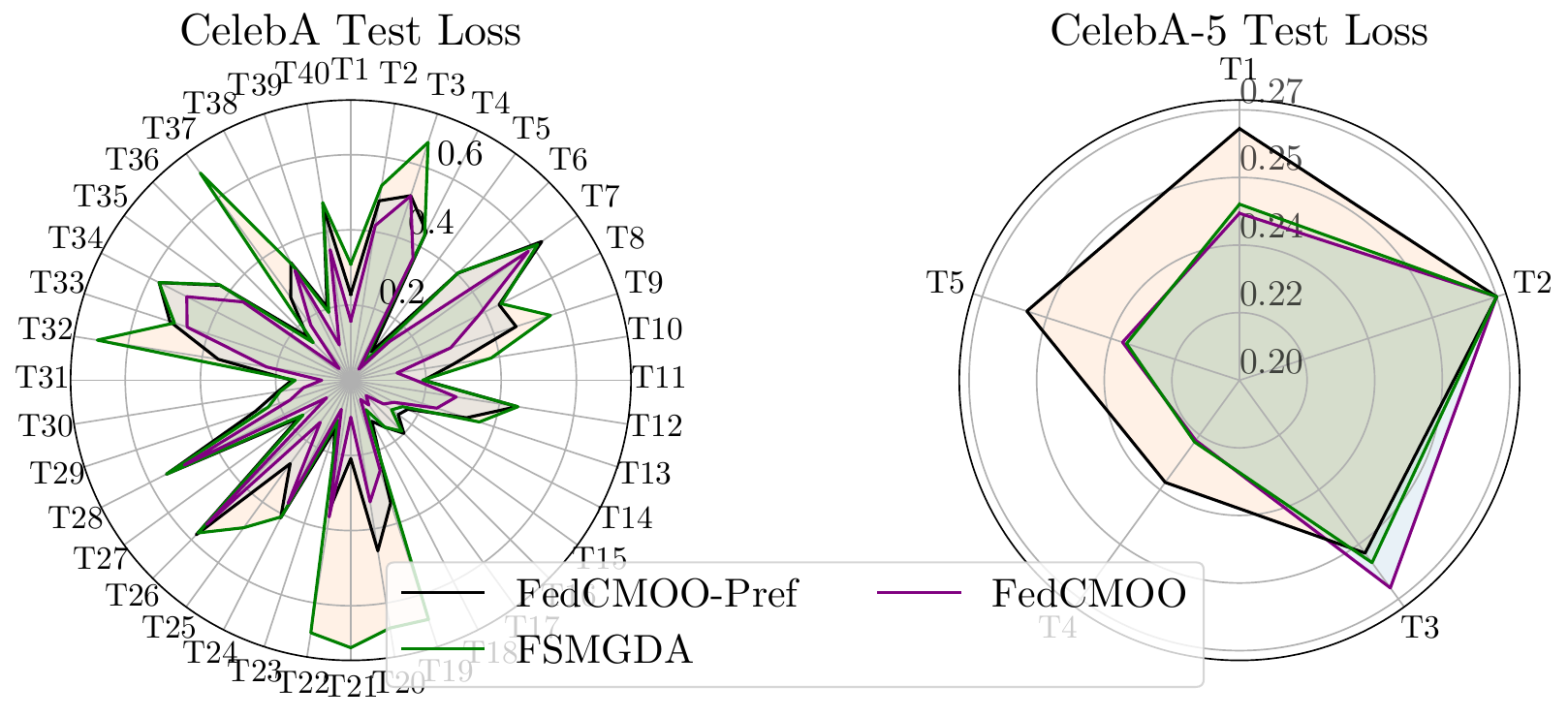}}
    \caption{Final test loss values of all objectives in the $\ce$ and $\cf$ experiments. $\fedmgda$ performs poorly as the number of objectives increases.}
    \label{fig:celebas_test_losses}
\end{figure}

We present the final loss values for each task in the $\ce$ and $\cf$ experiments shown in \Cref{fig:all_lr_loss_test_mean_all}, using radial plots in \Cref{fig:celebas_test_losses}. The performance of $\fedmgda$ worsens relative to other methods as the number of objectives increases due to local training drift. 

\subsection{\camera{$\Delta_M$ Metric Results of the Main Experiments in Table~\ref{tab:loss_L}}}\label{app_sect:delta_m}

\camera{$\Delta_M$ is a widely used metric {in MOO literature \citep{fernando2022mitigating}, measuring the average percentage performance loss} of multi-objective training compared to single-objective training. It is defined for a method $\mathcal{A}$ relative to a baseline $\mathcal{B}$ as:
\[
\Delta_M = \frac{1}{\M}\sum_{m=1}^\M (-1)^{\ell_m} \frac{S_{\mathcal{A},m} - S_{\mathcal{B},m}}{S_{\mathcal{B},m}},
\]
where $S_{\mathcal{A},m}$ and $S_{\mathcal{B},m}$ denote the performance of objective $m$ under methods $\mathcal{A}$ and $\mathcal{B}$, respectively. $\M$ is the number of objectives, and $\ell_m$ is $0$ if lower values of $S$ indicate better performance and $1$ otherwise \citep{fernando2022mitigating}.
\\
In Table~\ref{app_table:deltam}, we report $\Delta_M$ for the accuracy results of the experiments shown Table~\ref{tab:loss_L}. In this calculation, $S_{\mathcal{A},m}$ corresponds to the accuracy of objective $m$ in multi-objective training with algorithm $\mathcal{A}$, while $S_{\mathcal{B},m}$ is the accuracy when training the objective individually. For example, $\Delta_M$ for $\our$ on the MultiMNIST dataset is computed as:
\begin{align}
    \frac{1}{2} \bigg( &\frac{\text{Single-task Digit 1 accuracy} - \text{$\our$ Digit 1 accuracy}}{\text{Single-task Digit 1 accuracy}} \nn \\ &\quad\quad\quad\quad\quad\quad\quad\quad\quad\quad+ \frac{\text{Single-task Digit 2 accuracy} - \text{$\our$ Digit 2 accuracy}}{\text{Single-task Digit 2 accuracy}} \bigg). \nn
\end{align}
}

\begin{table}[h]
\centering
\caption{\camera{The $\Delta_M$ values for the test accuracy levels in experiments in Table~\ref{tab:loss_L} in the main text. Smaller values are better.}}
\label{app_table:deltam}
\begin{tabular}{c|c|c|c}
Experiment & \begin{tabular}[c]{@{}c@{}}MNIST+\\ FMNIST\end{tabular} & \begin{tabular}[c]{@{}c@{}}Multi\\ MNIST\end{tabular} & \begin{tabular}[c]{@{}c@{}}CIFAR10+\\ MNIST\end{tabular} \\ \hline
$\fedmgda$ \citep{FMGDA} & 10.22\% & $4.26\%$ & $5.63\%$ \\
$\ouri$ & $\mathbf{7.04\%}$ & $\mathbf{0.79\%}$ & $5.93\%$ \\
$\ourp$ & $7.60\%$ & $1.91\%$ & $\mathbf{5.12\%}$
\end{tabular}
\end{table}

\camera{While this metric reflects average performance, Table~\ref{tab:loss_L} in the main text provides detailed task-specific results. The highest achievable accuracies for the two-objective federated settings in Table~\ref{app_table:deltam} are as follows: For {$\mf$}, $97.0\%$ (MNIST) and $90.1\%$ (FashionMNIST); for {$\mm$}, $95.4\%$ (Digit 1) and $93.1\%$ (Digit 2); and for {$\cm$}, $64.9\%$ (CIFAR-10) and $97.1\%$ (MNIST).}

\subsection{Further Experiments Comparing The Local Training Performance Of \texorpdfstring{$\ournormal$}{FedCMOO} And \texorpdfstring{$\fedmgdanormal$}{FSMGDA}} \label{app_sect:local_progress_gap}

\begin{figure}[H]
\centerline{\includegraphics[width=0.75\textwidth]{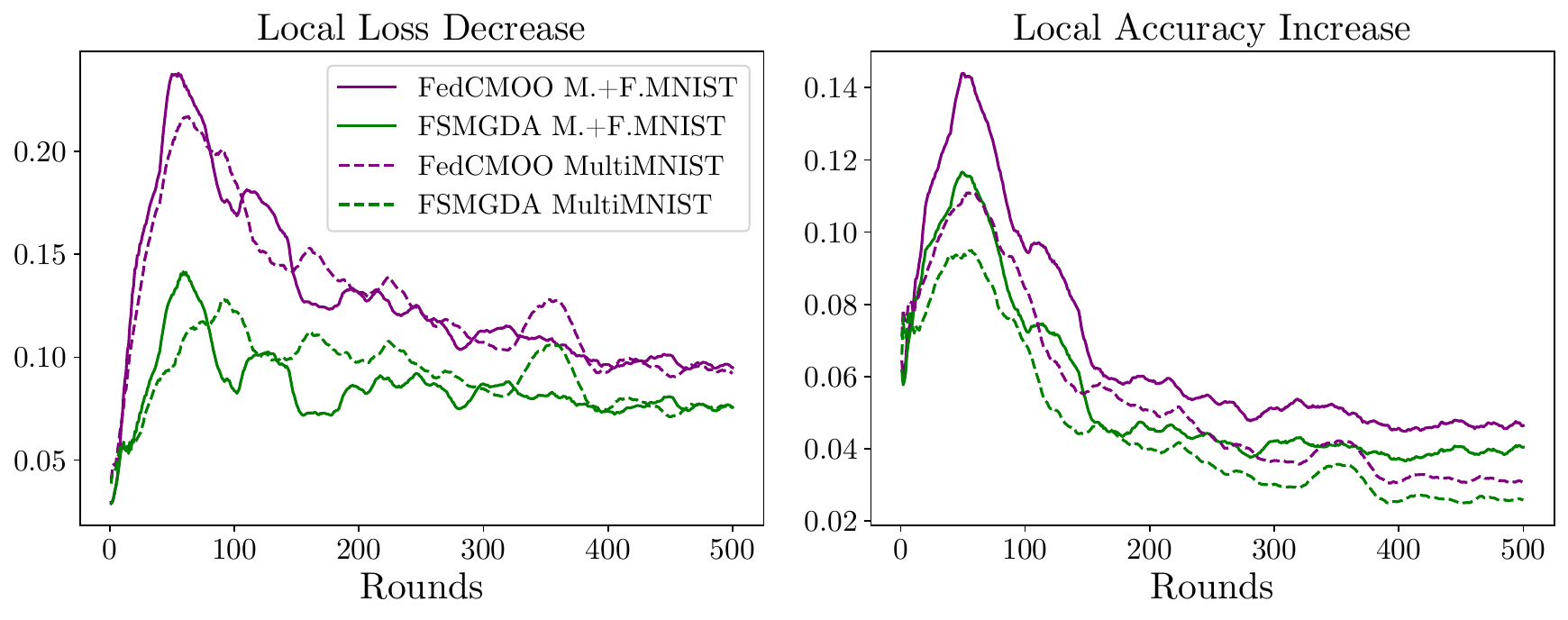}}
    \caption{Comparison of average local training progress (loss $\downarrow$ and accuracy $\uparrow$) across clients between $\our$ and $\fedmgda$. Drifting gradients during local training lead to poor performance in $\fedmgda$. For better visualization, the results are smoothed with a running average over $40$ rounds.}
    \label{fig:local_improvements}
\end{figure}

To further support our claim that local drift across objectives reduces the performance of $\fedmgda$, we designed the following experiment. During $\our$ training in the $\mm$ and $\mf$ experiments, we also evaluate the local training performance of $\fedmgda$ using the same global models. We assess the clients’ local training performance with both $\our$ and $\fedmgda$ by measuring accuracy and loss on local datasets at the beginning and end of each local training session. The difference in these values gives us the loss reduction and accuracy improvement during local training. These results are averaged over clients and plotted across rounds. The results, shown in \Cref{fig:local_improvements}, indicate that clients achieve greater local training progress with $\our$ than with $\fedmgda$. Drift across objectives in $\fedmgda$ hinders local training performance.

\subsection{\camera{The Final Test Loss Values of All Tasks in the QM9 Experiment}\label{app_sect:qm9_exp_res}}

\begin{figure}[H]
\centerline{\includegraphics[width=0.38\textwidth]{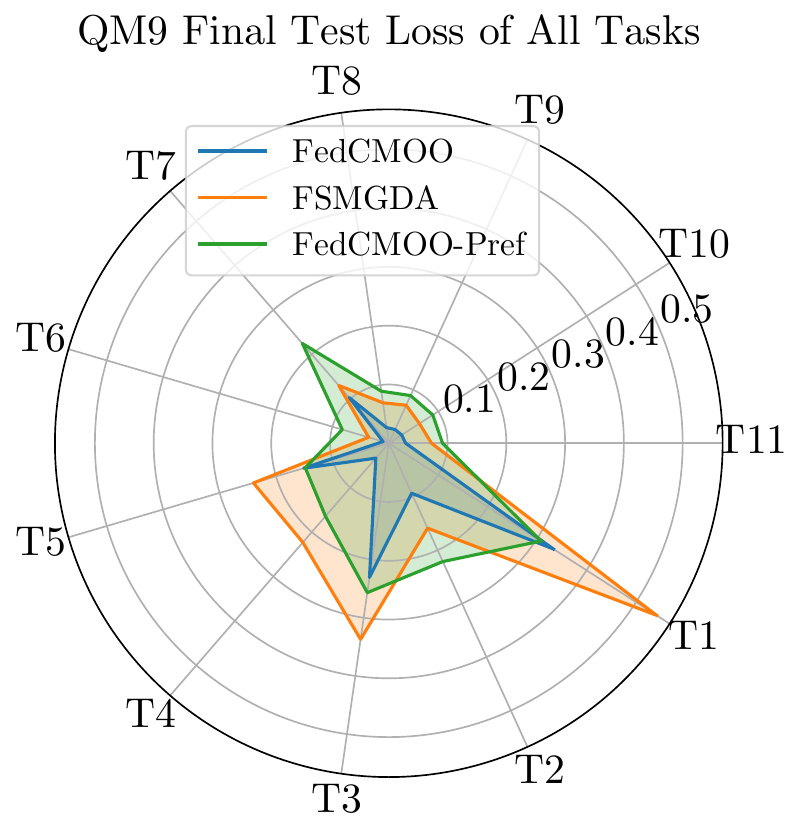}}
    \caption{\camera{Radial plot showing the final test loss of $\our$, $\ourp$ (with a uniform preference), and $\fedmgda$ for all tasks in the QM9 experiment. $\our$ achieves superior performance across tasks.}}
    \label{fig:qm9_radial}
\end{figure}

\camera{We present the radial plot of the final test accuracies of all individual tasks in $\qm$ experiment for each method in \Cref{fig:qm9_radial}. It further demonstrates the superiority of $\our$ over the baseline on every task.}

\section{\camera{THE IMPRACTICAL ASSUMPTION IN THE THEORETICAL ANALYSIS OF \cite{FMGDA}}}\label{app_sect:unusual_assump_in_yang_et_al}

\camera{In this section, we provide a detailed explanation expanding upon the discussion at the end of \Cref{sect:theory}, focusing on the nonstandard assumption in \cite{FMGDA}. We specifically refer to Assumption 4, Theorem 5, and Corollary 6, all of which are from \cite{FMGDA}. 

Assumption 4 in \cite{FMGDA} combines Lipschitz continuity and bounded gradient variance by assuming 
\(
\mathbb{E}\left[||\nabla f(\mathbf{x}, \xi) - \nabla f(\mathbf{y}, \xi’)||^2\right] \leq \alpha ||\mathbf{x} - \mathbf{y}||^2 + \beta \sigma^2.
\)
This assumption is reasonable, as there always exists a sufficiently large $\alpha$ and $\beta$. However, in Theorem 5, their bound on $\mathbb{E}||\bar{\mathbf{d}}_t||$ contains a $\mathcal{O}(\beta\sigma^2)$ term. From Assumption 4, this term cannot always be arbitrarily small. To ensure its convergence to zero, Corollary 6 imposes the assumption $\beta = \mathcal{O}(\eta)$. This is unusual, as $\beta$ is a constant that depends on the data distribution and is not a tunable parameter. Since $\eta = \mathcal{O}(1/\sqrt{T})$ is chosen, this assumption effectively forces the $\beta \sigma^2$ term in Assumption 4 to be $\mathcal{O}(\sigma^2/\sqrt{T})$, which is a very strong requirement. One way to satisfy this condition would be to make the batch size at each step depend on $T$. However, this would significantly worsen the convergence rate with respect to the number of data samples.
}
\section{PROOF OF CONVERGENCE THEOREM} \label{app_sect:proof}

We provide the mathematical proofs of the claims in the paper in this section.

\subsection{Definitions}

The norms are Frobenius norm unless otherwise specified. $\opnorm{A}$ denotes the operator norm of matrix $A$. The notation $\rnd$ with subscript is used to clarify the source of randomness when multiple sources could cause confusion.

\subsection{ \texorpdfstring{$\approxg$ and $\our$ Algorithm with All Simplifications for Theoretical Analysis}{ApproxGramJacobian}}\label{app_sect:changes_in_alg_theory}
As explained in \Cref{sect:theory} of the main text, we prove the convergence of $\our$ with slight modifications to $\approxg$ subroutine. The resulting algorithm, with the incorporated changes, is presented in \Cref{alg:approxg_for_proof}. This algorithm takes as input the model at round $t$, $\x^\tp{t}$, and returns an unbiased approximation of the Gram matrix of the Jacobian of the global loss vector, $\G\lvect\p{\x^\tp{t}}^\top\G\lvect\p{\x^\tp{t}}$. First, two independent sets of $\np$ clients are sampled. The clients in each set calculate local stochastic Jacobian matrices, compressing the columns (i.e., individual objectives’ stochastic gradients) with an unbiased compression operator $\Q$ before sending them to the server. On the server, the collected stochastic Jacobian matrices from both sets of clients are averaged. Using these two averaged matrices, the subroutine returns an unbiased estimate of the Gram matrix of the Jacobian of the global loss vector.

\begin{algorithm}
\caption{$\approxg$\texttt{-T}: $\approxg$ with incorporated changes for theoretical analysis}
\label{alg:approxg_for_proof}
\begin{algorithmic}[1]
\STATE \textbf{Input:} model $\x^\tp{t}$
    \STATE $\A_1^\tp{t}\gets$Sample $\np$ clients uniformly at random
    \STATE $\A_2^\tp{t}\gets$Sample $\np$ clients uniformly at random
    \FOR{$j\in\{1,2\}$ in parallel}
        \FOR{each client $\ii \in \A_j^\tp{t}$ in parallel}
            \STATE $\hh_\ii\p{\x^\tp{t},\rnd_j} \gets \sqbr{\sG\f_{\ii,1}\p{\x^\tp{t},\rnd_j},\dots,\sG\f_{\ii,\M}\p{\x^\tp{t},\rnd_j}} \in \mathbb{R}^{\xdim\times\M}$
            \STATE Compress each column of $\hh_\ii\p{\x^\tp{t},\rnd_j}$ with an unbiased compression operator $\Q$, and send $\Qp{\hh_\ii\p{\x^\tp{t},\rnd_j}}=\sqbr{\Qp{\sG\f_{\ii,1}\p{\x^\tp{t},\rnd_j}},\dots,\Qp{\sG\f_{\ii,\M}\p{\x^\tp{t},\rnd_j}}}$ to the server
        \ENDFOR
        \STATE $\yxtrnd{j}\gets \frac{1}{\np}\sum_{\ii\in \A_j^\tp{t}}\Qp{\hh_\ii\p{\x^\tp{t},\rnd_j}}$ at the server
    \ENDFOR
    \STATE \textbf{Return} $\yxtrnd{1}^\top\yxtrnd{2}$ 
\end{algorithmic}
\end{algorithm}

We will use \(\yxtrnd{j}\), as defined in Algorithm~\ref{alg:approxg_for_proof}, throughout the theoretical analysis. For the sake of completeness, we also present the full algorithm, for which the convergence guarantee is proven, in \Cref{alg:our_theory_app}.

\begin{algorithm}[H]
\caption{$\our$ with $\approxg$\texttt{-T} whose convergence guarantee is shown}
\label{alg:our_theory_app}
\begin{algorithmic}[1]
\STATE \textbf{Input:} client and server learning rates $\lrl, \lrg$, number of local steps $\locit$
\STATE \textbf{Initialize:} global model $\x^\tp{0} \in \mathbb{R}^\xdim$, and task weights $\w^\tp{0}\gets [1/\M,\dots,1/\M]^\top \in \WM$
\FOR{$t = 0, 1, \dots, T - 1$}
    \STATE Select the client set $\B^\tp{t}$ uniformly at random from $[N]$; send $\x^\tp{t}$ to the clients in $\B^\tp{t}$ 
    \STATE $G^\tp{t}\triangleq \yxtrnd{1}^\top\yxtrnd{2} \gets \approxg\texttt{-T} (\x^\tp{t})$ at the server, where \mbox{$G^{(t)} \approx \G \lvect (\x^{(t)})^\top \G \lvect (\x^{(t)})$} 
    \STATE Compute $\w^\tp{t+1} \gets \getw$($\w^\tp{t}, G^\tp{t}$, $K=1$)\label{line_app_fedcmoo:w_update}  \hfill \# $\w^\tp{t+1}\gets\projwm\p{\w^\tp{t}-\beta G^\tp{t}\w^\tp{t}}$\\at server and send $\w^\tp{t+1}$ to clients in $\B^\tp{t}$ 
    \FOR{each client $\ii \in \B^\tp{t}$ \textbf{in parallel}}
        \STATE Initialize local model: $\x_\ii^\tp{t,0} \gets \x^\tp{t}$ 
        \FOR{$r = 0, \dots, \locit-1$}
            \STATE 
            \(
            \x_\ii^\tp{t,r+1} \gets \x_\ii^\tp{t,r} - \lrl \sum_{k=1}^\M \wk^\tp{t+1} \sG\fik (\x_\ii^\tp{t,r})
            \) 
        \ENDFOR 
        \STATE Send
        \(
        \D_\ii^\tp{t} \triangleq \mfrac{1}{\locit \lrl} \big( \x^\tp{t} - \x_\ii^\tp{t,\locit} \big)
        \) to the server 
    \ENDFOR
    \STATE
    \(
    \x^\tp{t+1} \gets \x^\tp{t} - \lrg \lrl \locit \mfrac{1}{|\B^\tp{t}|} \sum_{\ii \in \B^\tp{t}} \D_\ii^\tp{t}
    \)
\ENDFOR
\STATE \textbf{Return} $\x^\tp{T}$ 
\end{algorithmic}
\end{algorithm}

\subsection{Restatement of Theoretical Claims}
We first restate the main theorem.

\begin{theorem}[Restatement of \Cref{thm:FedCMO}: Convergence of $\ourg$]
\label{app_thm:FedCMO}
Suppose Assumptions~\ref{assum:smoothness}-\ref{assum:compress} hold, the client learning rate satisfies $\lrl\leq \frac{1}{2\smo\locit}$, and the server selects $|\B^{(t)}| = \n$ clients in every round. Then the iterates of $\our$ (\Cref{alg:our_theory_app}) satisfy,
\begin{align}
     \frac{1}{T} \sum_{t=0}^{T-1} \E \norm{\sum_\kk \wk^\tp{t} \G \tl_k (\x^{(t)})}^2 \leq &\underbrace{\bigo{\beta\M\cone^2}}_{\textit{MOO Weight Error}}  + \underbrace{\bigo{\frac{1}{T\lrg \lrl \locit} + \frac{\smo\lrg\lrl\lhetsq}{\n}}}_{\substack{\textit{Centralized Optimization Error} \\ \textit{for Scalarized Loss}}} \nn \\ &+ \underbrace{\bigo{\smo\lrg\lrl\locit\bg^2}}_{\substack{\textit{Partial Participation} \\ \textit{Error}}} + \underbrace{\bigo{\smo \lrl (\locit \bg^2 + \sqrt{\locit}\bg\lhet)}}_{\textit{Local Drift Error}}, \nn
\end{align}
where \mbox{$\cone\triangleq \mco \p{\mfrac{\q+1}{\np}\lhetsq +\mfrac{\q\bg^2}{\np}+\mfrac{\N-\np}{\np(N-1)}\ghetsq + \bg^2}$} is a constant independent of $T$ and $\M$.
\end{theorem}

Also, by noticing that \mbox{$ \normsq{\sum_\kk \wk^{*\tp{t}} \G \tl_k (\x^{(t)})} \leq \normsq{\sum_\kk \wk^\tp{t} \G \tl_k (\x^{(t)})}$} for all $t$ and $\Jxt$ where \mbox{$\w^{*\tp{t}}\triangleq \arg\min_{\w\in\WM}\normsq{ \sum_\kk \wk \G \tl_k (\x^{(t)}) }$}, the following holds:
\begin{align}
    & \frac{1}{T} \sum_{t=0}^{T-1} \E \norm{\sum_\kk \wk^{*\tp{t}} \G \tl_k (\x^{(t)})}^2 \leq {\bigo{\beta\M\cone^2}}  + {\bigo{\frac{1}{T\lrg \lrl \locit} + \frac{\smo\lrg\lrl\lhetsq}{\n}}} + {\bigo{\smo\lrg\lrl\locit\bg^2}} + \bigo{\smo \lrl (\locit \bg^2 + \sqrt{\locit}\bg\lhet)}.\nn
\end{align}

\begin{cor}[Restatement of Corollary~\ref{cor:FedCMOO}: Convergence Rate]
\label{cor_app:fedcmoo}
With the client and server learning rates \mbox{$\lrl=\frac{1}{\smo\locit\sqrt{\locit T}}$}, \mbox{$\lrg=\sqrt{\locit}$}, and step size of \mbox{$\getw$ $\beta = \frac{1}{\M \sqrt{T}}$}, the bound on $\frac{1}{T}\sum_{t=0}^{T-1} \E \norm{\sum_\kk \wk^\tp{t} \G \tl_k}^2$ in (\ref{eq:theorem1}) reduces to,
\begin{align}
    \frac{1}{T}\sum_{t=0}^{T-1}\E\normsq{\Jxt\wt} \leq \bigo{\frac{\bg^2 + \cone^2}{\sqrt{T}}} + \bigo{\frac{\lhetsq}{\locit\sqrt{T}} \p{\frac{1}{\n} + \frac{1}{\locit}}}. \nn
\end{align}
\end{cor}

\subsection{Intermediate Lemmas}\label{app_sect:intermediate_lemmas}
We first provide the intermediate lemmas used throughout the proof. The expectation used in this section is expectation conditioned on $\x^\tp{t}$ and $\wt$. We drop the condition notation for brevity.

\begin{lemma}[Linear Algebra Tools] \label{lemma0}
 For a matrix $A\in\mathbb{R}^{\xdim\times\M}$ with columns $\mathbf{a}_1,\dots,\mathbf{a}_M$ such that $\norm{\mathbf{a}_\kk}\leq\bg$ for all $\kk\in[\M]$, and a vector $\w\in\WM$ where $\WM$ is $\M$-probability simplex, the followings hold.
\begin{enumerate}
    \item $||A\w||\leq\bg$. \textit{Proof}: $||A\w||=||\sum_k\wk \mathbf{a}_k||\leq \sum_k \wk ||\mathbf{a}_k||\leq \sum_k \wk\bg=\bg$.
    \item $\opnorm{A}\leq\bg\sqrt{\M}$. \textit{Proof:} $\opnorm{A}\leq||A||_2=\sqrt{\sum_k||\mathbf{a}_k||^2}\leq\sqrt{\sum_k \bg^2}=\bg\sqrt{\M}$.
    \item $\norm{A^\top A\w}\leq\sqrt{\M}\bg^2$. \textit{Proof:} $\norm{A^\top A\w}\leq\opnorm{A}\norm{A\w}\leq\bg^2\sqrt{M}$ using the first two inequalities.
\end{enumerate}
\end{lemma}

\begin{lemma}\label{lemma1a}
Suppose Assumptions~\ref{assum:lochet}~-~\ref{assum:compress} hold. Then, the expected value norm difference of averaged compressed stochastic Jacobian matrices across clients and exact Jacobian is bounded. Recall that \(\yxtrnd{j}\) is defined in Algorithm~\ref{alg:approxg_for_proof} as the averaged compressed stochastic Jacobian estimates from sampled clients.
\begin{align*}
\E\norm{\yxtrnd{j}-\Jxt}^2&\leq  \M\p{ \frac{1+q}{\np} + \frac{q}{\np}\bg^2 + \frac{\N-\np}{\np(\N-1)}\ghetsq },\\
&\text{and,} \\
\E\opnorm{\yxtrnd{j}-\Jxt}&\leq\sqrt{\M}\sqrt{\frac{\q+1}{\np}\lhetsq+\frac{\q\bg^2}{\np}+\frac{\N-\np}{\np(\N-1)}\ghetsq}.
\end{align*}

The proof is presented in \Cref{app_sec:proofs_of_intermediate_lemmas}.
\end{lemma}

\begin{lemma}\label{lemma1b}
Suppose Assumptions~\ref{assum:lochet}~-~\ref{assum:compress} hold. Then, the expected value norm difference of the weighted average of averaged compressed stochastic gradients across clients and exact gradients is bounded. 
\begin{align*}
\E\norm{ \p{\yxtrnd{j}-\Jxt}\wt }&\leq \sqrt{\frac{\q+1}{\np}\lhetsq+\frac{\q\bg^2}{\np}+\frac{\N-\np}{\np(\N-1)}\ghetsq},\\
&\text{and,} \\
\E\normsq{ \p{\yxtrnd{j}-\Jxt}\wt }&\leq  \frac{1+q}{\np}\lhetsq + \frac{q}{\np}\bg^2 + \frac{\N-\np}{\np(\N-1)}\ghetsq .
\end{align*}

The proof is presented in \Cref{app_sec:proofs_of_intermediate_lemmas}.
\end{lemma}

\begin{lemma}\label{lemma1c}
Suppose Assumptions~\ref{assum:lochet}~-~\ref{assum:compress} hold. Then, the update on task weights (line~\ref{line_app_fedcmoo:w_update} in \Cref{alg:our_theory_app}) is bounded.
\allowdisplaybreaks{\begin{align}
    \E\norm{\wt-\wtp}\leq\beta\sqrt{\M}\cone,
\end{align}}

where \(\cone\triangleq \p{\sqrt{\frac{1+q}{\np} + \frac{q}{\np}\bg^2 + \frac{\N-\np}{\np(\N-1)}\ghetsq}+\bg}^2\) is a constant independent of $T$ and $M$. Also, note that \mbox{$\cone\leq\bigo{\frac{1+q}{\np} + \frac{q}{\np}\bg^2 + \frac{\N-\np}{\np(\N-1)}\ghetsq+\bg^2}$} by AM-GM inequality.

The proof is presented in \Cref{app_sec:proofs_of_intermediate_lemmas}.
\end{lemma}

\begin{lemma}\label{lemma2}
    Suppose Assumptions~\ref{assum:lochet}~-~\ref{assum:compress} hold. Then we can show the following inequality,
\allowdisplaybreaks{\begin{align}
    \inp{\Jxt\wt-\Jxt\w}{\Jxt\wt} \leq \frac{\E\normsq{\wt-\w} - \E\normsq{\wtp-\w}}{2\beta}+\frac{\beta\M\cone^2}{2}, \nn
\end{align}}

where \(\cone\triangleq \p{\sqrt{\frac{1+q}{\np} + \frac{q}{\np}\bg^2 + \frac{\N-\np}{\np(\N-1)}\ghetsq}+\bg}^2\) is a constant independent of $T$ and $M$.

The proof is presented in \Cref{app_sec:proofs_of_intermediate_lemmas}.
\end{lemma}

\begin{lemma} \label{lemma3}
    Suppose Assumptions~\ref{assum:lochet}~ and ~\ref{assum:bounded_grad} hold. The aggregated update of clients is bounded in expectation such that,
    \allowdisplaybreaks{\begin{align}
        \E\normsq{\frac{1}{\n}\sum_{\ii\in\B^\tp{t}}\Dit}\leq\bg^2+\frac{\lhetsq}{\n\locit}.
    \end{align}}

The proof is presented in \Cref{app_sec:proofs_of_intermediate_lemmas}.
\end{lemma}

\begin{lemma} \label{lemma4}
    Suppose Assumptions~\ref{assum:smoothness}, \ref{assum:lochet} and \ref{assum:bounded_grad} hold, and $\lrl\leq\frac{1}{2\smo\locit}$. Then, we can show the following inequality,
    \allowdisplaybreaks{\begin{align}
        \E\sqbr{\inp{\Jxt\w}{-\p{\frac{1}{\n}\sum_{\ii\in\B^\tp{t}} \Dit -\Jxt\wtp}}} \leq \smo\lrl\sqrt{\locit}\lhet\bg + \smo\lrl\bg^2\sqrt{2\locit(\locit-1)}.
    \end{align}}
    
The proof is presented in \Cref{app_sec:proofs_of_intermediate_lemmas}.
\end{lemma}

\subsection{Proofs of Main Statements}
We present the proofs of \Cref{thm:FedCMO} and Corollary~\ref{cor:FedCMOO}. We are inspired by some proof techniques in \cite{direction_oriented, fedvarp, fedast} in lemmas and main proof. 

\subsubsection{Proof of \Cref{thm:FedCMO}}
The expectation used in this section is expectation conditioned on $\x^\tp{t}$ and $\wt$. We drop the condition notation for brevity. Let us fix some $\w\in\WM$. Using the update rule in \Cref{alg:our} and $\asmp$~\ref{assum:lochet},
\allowdisplaybreaks{\begin{align}
    & \lvect\pxtp\w \leq \lvect\pxt\w + \lrl\lrg\locit\inp{\Jxt\w}{-\fsumc\Dit} + \frac{\smo\lrl^2\lrg^2\locit^2}{2}\normsq{\fsumc\Dit} \nn \\
    & = \lvect\pxt\w + \lrl\lrg\locit\inp{\Jxt\w}{-\p{\fsumc\Dit-\Jxt\wtp}} \nn\\
    & \quad + \frac{\smo\lrl^2\lrg^2\locit^2}{2}\normsq{\fsumc\Dit} + \lrl\lrg\locit\inp{\Jxt\w}{-\Jxt\wtp}. \nn
\end{align}}

Then, taking conditional expectation on $\xt$ and $\wt$,
\allowdisplaybreaks{\begin{align}
& \E\sqbr{\lvect\pxtp\w}-\lvect\pxt\w \leq \lrl\lrg\locit\E\sqbr{\inp{\Jxt\w}{-\p{\fsumc\Dit-\Jxt\wtp}}}\nn\\
    & \quad + \frac{\smo\lrl^2\lrg^2\locit^2}{2}\E\normsq{\fsumc\Dit} + \lrl\lrg\locit\inp{\Jxt\w}{-\Jxt\E\sqbr{\wtp}} \nn \\
    & = \lrl\lrg\locit\E\sqbr{\inp{\Jxt\w}{-\p{\fsumc\Dit-\Jxt\wtp}}} + \frac{\smo\lrl^2\lrg^2\locit^2}{2}\E\normsq{\fsumc\Dit} \nn \\
    &  + \lrl\lrg\locit\inp{\Jxt\w}{\Jxt\E\sqbr{\wt-\wtp}} + \lrl\lrg\locit\inp{\Jxt\w}{-\Jxt\wt} \nn \\
    & = \lrl\lrg\locit\E\sqbr{\inp{\Jxt\w}{-\p{\fsumc\Dit-\Jxt\wtp}}} + \frac{\smo\lrl^2\lrg^2\locit^2}{2}\E\normsq{\fsumc\Dit} \nn \\
    &  + \lrl\lrg\locit\inp{\Jxt\w}{\Jxt\E\sqbr{\wt-\wtp}} \mp \lrl\lrg\locit\normsq{\Jxt\wt} \nn \\
    & + \lrl\lrg\locit\inp{\Jxt\w}{-\Jxt\wt}  \nn \\
    & = \lrl\lrg\locit\E\sqbr{\inp{\Jxt\w}{-\p{\fsumc\Dit-\Jxt\wtp}}} + \frac{\smo\lrl^2\lrg^2\locit^2}{2}\E\normsq{\fsumc\Dit} \nn \\
    &  + \lrl\lrg\locit\E\sqbr{\inp{\Jxt^\top\Jxt\w}{\wt-\wtp}} - \lrl\lrg\locit\normsq{\Jxt\wt}\nn \\
    & + \lrl\lrg\locit\inp{\Jxt\wt-\Jxt\w}{\Jxt\wt}  \nn \\
    &\leqos{\\\text{Cauch-Schwarz}\\\text{inequality}} \lrl\lrg\locit\E\sqbr{\inp{\Jxt\w}{-\p{\fsumc\Dit-\Jxt\wtp}}} + \frac{\smo\lrl^2\lrg^2\locit^2}{2}\E\normsq{\fsumc\Dit} \nn \\
    & +\lrl\lrg\locit\norm{\Jxt^\top\Jxt\w}\E\norm{\wt-\wtp} - \lrl\lrg\locit\normsq{\Jxt\wt}\nn \\
    & + \lrl\lrg\locit\inp{\Jxt\wt-\Jxt\w}{\Jxt\wt}.  \nn
\end{align}}

Then using Lemmas \ref{lemma0} and \ref{lemma1c}, and defining \(\cone\triangleq \p{\sqrt{\frac{1+q}{\np} + \frac{q}{\np}\bg^2 + \frac{\N-\np}{\np(\N-1)}\ghetsq}+\bg}^2\),
\allowdisplaybreaks{\begin{align}
& \E\sqbr{\lvect\pxtp\w}-\lvect\pxt\w \leq \lrl\lrg\locit\E\sqbr{\inp{\Jxt\w}{-\p{\fsumc\Dit-\Jxt\wtp}}}  \nn \\
& + \frac{\smo\lrl^2\lrg^2\locit^2}{2}\E\normsq{\fsumc\Dit} + \lrl\lrg\locit\beta\M\cone\bg^2 + \lrl\lrg\locit\inp{\Jxt\wt-\Jxt\w}{\Jxt\wt} \nn \\
& - \lrl\lrg\locit\normsq{\Jxt\wt} \nn \\
& \leqos{\\\text{\Cref{lemma2}}} \lrl\lrg\locit\E\sqbr{\inp{\Jxt\w}{-\p{\fsumc\Dit-\Jxt\wtp}}} + \frac{\smo\lrl^2\lrg^2\locit^2}{2}\E\normsq{\fsumc\Dit} \nn \\
&  + \lrl\lrg\locit\beta\M\cone\bg^2 +\lrl\lrg\locit\frac{\normsq{\wt-\w}-\E\normsq{\wtp-\w}}{2\beta}  + \lrl\lrg\locit\frac{\beta\M\cone^2}{2} - \lrl\lrg\locit\normsq{\Jxt\wt}  \nn \\
& \leqos{\\\text{\Cref{lemma3} and}\\\text{assuming }\lrl\leq\tfrac{1}{2\smo\locit}} \lrl\lrg\locit\E\sqbr{\inp{\Jxt\w}{-\p{\fsumc\Dit-\Jxt\wtp}}} - \lrl\lrg\locit\normsq{\Jxt\wt} \nn \\
& + \lrl\lrg\locit\beta\M\p{\cone\bg^2+\frac{\cone^2}{2}} +\lrl\lrg\locit\frac{\normsq{\wt-\w}-\E\normsq{\wtp-\w}}{2\beta}  + \frac{\smo\lrl^2\lrg^2\locit^2}{2}\p{\bg^2+\frac{\lhetsq}{\n\locit}} \nn \\
& \leqos{\\\text{\Cref{lemma4}}} \lrl\lrg\locit\p{\smo\lrl\sqrt{\locit}\lhet\bg + \smo\lrl\bg^2\sqrt{2\locit(\locit-1)}} - \lrl\lrg\locit\normsq{\Jxt\wt} \nn \\
& + \lrl\lrg\locit\beta\M\p{\cone\bg^2+\frac{\cone^2}{2}} +\lrl\lrg\locit\frac{\normsq{\wt-\w}-\E\normsq{\wtp-\w}}{2\beta}  + \frac{\smo\lrl^2\lrg^2\locit^2}{2}\p{\bg^2+\frac{\lhetsq}{\n\locit}}. \nn
\end{align}}

Arranging the terms, we have
\allowdisplaybreaks{\begin{align}
    & \normsq{\Jxt\wt} \leq \frac{\lvect\pxt\w - \E\sqbr{\lvect\pxtp\w}}{\lrl\lrg\locit} + \frac{\normsq{\wt-\w}-\E\normsq{\wtp-\w}}{2\beta} + \beta\M\p{\cone\bg^2+\frac{\cone^2}{2}}  \nn \\
    & + \smo\lrl\sqrt{\locit}\lhet\bg + \smo\lrl\bg^2\sqrt{2\locit(\locit-1)} + \frac{\smo\lrl\lrg\locit}{2}\p{\bg^2+\frac{\lhetsq}{\n\locit}} \nn
\end{align}}

We choose $\w=\w^\tp{0}$. Define $\delta=\sum_{\kk\in[\M]}\wk^\tp{0}\p{\tl_\kk\p{\x^\tp{0}} - \min_\x\tl_\kk\p{\x}}$ Telescoping the terms through \mbox{$t=0,\dots,T-1$}, and taking the unconditional expectation by using the tower property,
\allowdisplaybreaks{\begin{align}
    &\frac{1}{T}\sum_{t=0}^{T-1}\E\normsq{\Jxt\wt} \leq \frac{\delta}{T\lrl\lrg\locit} + \beta\M\p{\cone\bg^2+\frac{\cone^2}{2}} + \smo\lrl\sqrt{\locit}\lhet\bg + \smo\lrl\bg^2\sqrt{2\locit(\locit-1)} + \frac{\smo\lrl\lrg\locit}{2}\p{\bg^2+\frac{\lhetsq}{\n\locit}} \nn \\
    & \leq \underbrace{\bigo{\beta\M\cone^2}}_{\textit{MOO Weight Error}}  + \underbrace{\bigo{\frac{1}{T\lrg \lrl \locit} + \frac{\smo\lrg\lrl\lhetsq}{\n}}}_{\substack{\textit{Centralized Optimization Error} \\ \textit{for Scalarized Loss}}} + \underbrace{\bigo{\smo\lrg\lrl\locit\bg^2}}_{\substack{\textit{Partial Participation} \\ \textit{Error}}} + \underbrace{\bigo{\smo \lrl (\locit \bg^2 + \sqrt{\locit}\bg\lhet)}}_{\textit{Local Drift Error}}, \label{line0:main_proof}
\end{align}}
where $\bigo{\cdot}$ swallows numerical constant dependencies and $\delta$ is the optimization gap of the initial model. 

\subsubsection{Proof of Corollary~\ref{cor:FedCMOO}}
Let $\bigo{\cdot}$ swallow smoothness constant $\smo$ as well for brevity. Now, inserting client and server learning rates \mbox{$\lrl=\frac{1}{\smo\locit\sqrt{\locit T}}$}, \mbox{$\lrg=\sqrt{\locit}$}, and step size of \mbox{$\getw$ $\beta = \frac{1}{\M \sqrt{T}}$} in \Cref{line0:main_proof}, we have
\allowdisplaybreaks{\begin{align}
    &\frac{1}{T}\sum_{t=0}^{T-1}\E\normsq{\Jxt\wt} \leq  {\bigo{\frac{\cone^2}{\sqrt{T}}}} + {\bigo{\frac{1}{\sqrt{T}} + \frac{\lhetsq}{\sqrt{T}\locit\n}}} + {\bigo{\frac{\bg^2}{\sqrt{T}}}} + {\bigo{\frac{\bg^2}{\sqrt{T\locit}} + \frac{\bg\lhet}{\locit\sqrt{T}}}} \nn \\
    & \eqos{\\\text{by AM-GM inequality}} \bigo{\frac{\bg^2 + \cone^2}{\sqrt{T}}} + \bigo{\frac{\lhetsq}{\locit\sqrt{T}} \p{\frac{1}{\n} + \frac{1}{\locit}}}. \nn
\end{align}}

\subsection{Proofs of Intermediate Lemmas} \label{app_sec:proofs_of_intermediate_lemmas}
Below, we provide the proofs of intermediate lemmas presented in \Cref{app_sect:intermediate_lemmas}.
\paragraph{Proof of \Cref{lemma1a}.}
To prove \Cref{lemma1a}, we first show that $\E\norm{\yxtrnd{j}-\Jxt}^2$ is bounded, which is the first part of the lemma.
\allowdisplaybreaks{\begin{align}
    & \E\norm{\yxtrnd{j}-\Jxt}^2  \nn \\
    &=\E\normsq{\sqbr{\frac{1}{\np}\sum_{\ii\in\A_j^\tp{t}}\p{\Qp{\sG\f_{\ii,1}\pxtrndj} - \G\tl_1\pxt},  \dots, \frac{1}{\np}\sum_{\ii\in\A_j^\tp{t}} \p{\Qp{\sG\f_{\ii,\M}\pxtrndj} - \G\tl_\M\pxt} } } \nn\\
    &= \sum_{\kk=1}^\M\E\normsq{\frac{1}{\np}\sum_{\ii\in\A_j^\tp{t}} \p{\Qp{\sG\fik\pxtrndj} - \G\tl_\kk\pxt} }\label{lineFor1b:lemma1a} \\
    &= \sum_{\kk=1}^\M\E\normsq{\frac{1}{\np}\sum_{\ii\in\A_j^\tp{t}} \p{\Qp{\sG\fik\pxtrndj} \mp \sG\fik\pxtrndj \mp \G\fik\pxtrndj - \G\tl_\kk\pxt } }\nn\\
    &= \sum_{\kk=1}^\M \E\sqbr{\frac{1}{(\np)^2}\sum_{\ii\in\A^\tp{t}_j}\p{\normsq{\Qp{\sG\fik\pxtrndj}-\sG\fik\pxtrndj} + \normsq{\sG\fik\pxtrndj - \G\fik\pxtrndj
    } }}\nn\\
    &\quad\quad\quad\quad\quad\quad\quad\quad\quad\quad\quad\quad\quad\quad\quad\quad\quad\quad\quad\quad\quad\quad\quad\quad+ \sum_{\kk=1}^\M\E\normsq{\frac{1}{\np}\sum_{\ii\in\A^\tp{t}_j}\p{\G\fik\pxt-\G\tl_\kk\pxt}}   \label{line1:lemma1a} \\
    &\leq\sum_{\kk=1}^\M \sqbr{\frac{1}{\np\N}\sum_{\ii\in[N]}\p{\q\E\normsq{\sG\fik\pxtrndj\mp\G\fik\pxt} + \lhetsq} + \E\normsq{\frac{1}{\np}\sum_{\ii\in\A^\tp{t}_j}\p{\G\fik\pxt-\G\tl_\kk\pxt}}   } \label{line2:lemma1a} \\
    &\leq \sum_{\kk=1}^\M \sqbr{\frac{1}{\np\N}\sum_{\ii\in[N]}\p{\q\E\normsq{\sG\fik\pxtrndj-\G\fik\pxt}+\q\normsq{\G\fik\pxt} + \lhetsq} + \frac{\N-\np}{\np(\N-1)}\ghetsq } \label{line3:lemma1a} \\
    &\leq \sum_{\kk=1}^\M \sqbr{\frac{1}{\np\N}\sum_{\ii\in[N]}\p{\q\lhetsq+\q\bg^2+ \lhetsq} + \frac{\N-\np}{\np(\N-1)}\ghetsq } \label{line4:lemma1a} \\
    &\leq  \M\p{ \frac{1+q}{\np}\lhetsq + \frac{q}{\np}\bg^2 + \frac{\N-\np}{\np(\N-1)}\ghetsq }, \label{line5:lemma1a}
\end{align}}

where \Cref{line1:lemma1a} follows observing that cross-terms will be zero in the expansion since the randomness across compression, stochastic gradients, and client selection is independent. \Cref{line2:lemma1a} holds by Assumptions~\ref{assum:lochet} and \ref{assum:compress}. \Cref{line3:lemma1a} follows the independence of stochastic gradients, Assumption~\ref{assum:globhet}, and Lemma 4 by \cite{fedvarp}. \Cref{line4:lemma1a} uses Assumptions~\ref{assum:lochet} and \ref{assum:bounded_grad}. This proves the first part of the lemma. Now,

\allowdisplaybreaks{\begin{align}
    &\E\opnorm{\yxtrnd{j}-\Jxt} \leq \E\norm{\yxtrnd{j}-\Jxt}_2 \nn \\
    &= \E\sqrt{\norm{\yxtrnd{j}-\Jxt}^2} \leqos{\\\textit{Jensen's}\\\textit{Inequality}} \sqrt{\E\norm{\yxtrnd{j}-\Jxt}^2} \nn \\
    &\leqos{\\\text{Using}\\\text{\Cref{line5:lemma1a}}} \sqrt{\M}\sqrt{\frac{\q+1}{\np}\lhetsq+\frac{\q\bg^2}{\np}+\frac{\N-\np}{\np(\N-1)}\ghetsq}. \nn
\end{align}}

\paragraph{Proof of \Cref{lemma1b}.}
To prove the lemma using the Assumptions~\ref{assum:lochet}~-~\ref{assum:compress}, we first observe that for any $\kk\in[\M]$,
\allowdisplaybreaks{\begin{align}
    \E\normsq{\frac{1}{\np}\sum_{\ii\in\A_j^\tp{t}} \p{\Qp{\sG\fik\pxtrndj} - \G\tl_\kk\pxt} } \leq  \frac{1+q}{\np}\lhetsq + \frac{q}{\np}\bg^2 + \frac{\N-\np}{\np(\N-1)}\ghetsq, \label{line0:lemma1b}
\end{align}}
following the same steps after \Cref{lineFor1b:lemma1a} in \Cref{lemma1a} without the outermost summation. Then,

\allowdisplaybreaks{\begin{align}
    & \E\norm{ \p{\yxtrnd{j}-\Jxt}\wt }\leqos{\\\text{Triangle}\\\text{inequality}} \sum_{\kk\in[\M]}\wkt\E\norm{\frac{1}{\np}\sum_{\ii\in\A_j^\tp{t}} \p{\Qp{\sG\fik\pxtrndj} - \G\tl_\kk\pxt} } \nn \\
    & = \sum_{\kk\in[\M]}\wkt\E\sqrt{ \normsq{\frac{1}{\np}\sum_{\ii\in\A_j^\tp{t}} \p{\Qp{\sG\fik\pxtrndj} - \G\tl_\kk\pxt} } } \nn \\
    & \leqos{\\\text{Jensen's}\\\text{inequality}} \sum_{\kk\in[\M]}\wkt\sqrt{ \E\normsq{\frac{1}{\np}\sum_{\ii\in\A_j^\tp{t}} \p{\Qp{\sG\fik\pxtrndj} - \G\tl_\kk\pxt} } } \nn\\
    & \leqos{\\\text{\Cref{line0:lemma1b}}} \sum_{\kk\in[\M]}\wkt\sqrt{ \frac{1+q}{\np}\lhetsq + \frac{q}{\np}\bg^2 + \frac{\N-\np}{\np(\N-1)}\ghetsq } \nn \\
    & \eqos{\\\sum_\kk\wkt=1} \sqrt{\frac{\q+1}{\np}\lhetsq+\frac{\q\bg^2}{\np}+\frac{\N-\np}{\np(\N-1)}\ghetsq}. \nn
\end{align}}
This proves the first part of the lemma. Then, we prove the second part,
\allowdisplaybreaks{\begin{align}
    &\E\normsq{ \p{\yxtrnd{j}-\Jxt}\wt } = \E\normsq{ \sum_{\kk\in[\M]}\wkt \p{\frac{1}{\np}\sum_{\ii\in\A_j^\tp{t}} \p{\Qp{\sG\fik\pxtrndj} - \G\tl_\kk\pxt}} } \nn\\
    & \leqos{\\\\\text{Jensen's}\\\text{inequality}} \sum_{\kk\in[\M]}\wkt \E\normsq{ {\frac{1}{\np}\sum_{\ii\in\A_j^\tp{t}} \p{\Qp{\sG\fik\pxtrndj} - \G\tl_\kk\pxt}} } \nn\\
    &\leqos{\\\text{\Cref{line0:lemma1b}}} \sum_{\kk\in[\M]}\wkt \p{ \frac{1+q}{\np}\lhetsq + \frac{q}{\np}\bg^2 + \frac{\N-\np}{\np(\N-1)}\ghetsq } \eqos{\\\sum_\kk\wkt=1} \frac{1+q}{\np}\lhetsq + \frac{q}{\np}\bg^2 + \frac{\N-\np}{\np(\N-1)}\ghetsq.\nn
\end{align}}

\paragraph{Proof of \Cref{lemma1c}.}
\allowdisplaybreaks{\begin{align}
    &\E\norm{\wt-\wtp}=\E\norm{\projwm\wt-\projwm\p{\wt-\beta\yxtrnd{1}^\top\yxtrnd{2}\wt}} \nn\\
    & \leq \E\norm{\wt-\p{\wt-\beta\yxtrnd{1}^\top\yxtrnd{2}\wt}} \label{line1:lemma1c} \\
    & = \beta\E\norm{\yxtrnd{1}^\top\yxtrnd{2}\wt} \nn \\
    & = \beta\E\norm{\p{\yxtrnd{1}\mp\Jxt}^\top\p{\yxtrnd{2}\mp\Jxt}\wt} \nn \\
    &\leqos{\\\text{Triangle}\\\text{inequality}}\beta\E\Bigg[ \norm{\p{\yxtrnd{1}-\Jxt}^\top\p{\yxtrnd{2}-\Jxt}\wt} + \norm{\Jxt^\top\Jxt\wt} \nn \\
    &+ \norm{\p{\yxtrnd{1}-\Jxt}^\top\Jxt\wt} + \norm{\Jxt^\top\p{\yxtrnd{2}-\Jxt}\wt} \Bigg]  \nn \\
    & \leqos{\\\text{Cauchy-Schwarz}\\\text{inequality}} \beta\Bigg[ \E\opnorm{\yxtrnd{1}-\Jxt}\E\norm{\p{\yxtrnd{2}-\Jxt}\wt}  \nn\\
    & + \E\opnorm{\Jxt}\E\norm{\Jxt\wt} + \E\opnorm{\yxtrnd{1}-\Jxt}\E\norm{\Jxt\wt} \nn\\
    & + \E\opnorm{\Jxt}\E\norm{\p{\yxtrnd{2}-\Jxt}\wt} \nn\\
    & \leqos{\\\Cref{lemma0}\\\Cref{lemma1a}\\\Cref{lemma1b}}\beta\sqrt{\M}\p{\frac{1+q}{\np}\lhetsq + \frac{q}{\np}\bg^2 + \frac{\N-\np}{\np(\N-1)}\ghetsq + 2\bg\sqrt{\frac{1+q}{\np}\lhetsq + \frac{q}{\np}\bg^2 + \frac{\N-\np}{\np(\N-1)}\ghetsq}+\bg^2} \nn \\
    & = \beta\sqrt{\M}\p{\sqrt{\frac{1+q}{\np}\lhetsq + \frac{q}{\np}\bg^2 + \frac{\N-\np}{\np(\N-1)}\ghetsq}+\bg}^2=\beta\sqrt{\M}\cone. \nn
\end{align}}
where \Cref{line1:lemma1c} follows the contraction property of the projection onto the convex sets, e.g., $\WM$, and \(\cone\triangleq \p{\sqrt{\frac{1+q}{\np}\lhetsq + \frac{q}{\np}\bg^2 + \frac{\N-\np}{\np(\N-1)}\ghetsq}+\bg}^2\).

\paragraph{Proof of \Cref{lemma2}.}
To prove the lemma, we first show $\E\normsq{\yxtrnd{1}^\top\yxtrnd{2}\wt}$ is bounded.
\allowdisplaybreaks{\begin{align}
&\E\normsq{\yxtrnd{1}^\top\yxtrnd{2}\wt}=\E\normsq{\p{\yxtrnd{1}\mp\Jxt}^\top\p{\yxtrnd{2}\mp\Jxt}\wt} \nn \\
&\leqos{\\\text{Cauchy-Schwarz and}\\\text{independence}\\\text{of $\rnd_1$ and $\rnd_2$}}\E\normsq{\yxtrnd{1}\mp\Jxt}\E\normsq{\p{\yxtrnd{2}\mp\Jxt}\wt} \nn \\
& =\p{\E\normsq{\yxtrnd{1}-\Jxt}+\E\normsq{\Jxt}} \Bigg( \E\normsq{\p{\yxtrnd{2}-\Jxt}\wt} \\&\hspace{33em}+\E\normsq{\Jxt\wt} \Bigg) \nn \\
&\leqos{\\\Cref{lemma0}\\\Cref{lemma1a}\\\Cref{lemma1b}}\M\p{\frac{\q+1}{\np}\lhetsq + \frac{\q\bg^2}{\np} + \frac{\N-\np}{\np(\N-1)}\ghetsq+ \bg^2}^2 \leq \M\cone^2, \label{line0:lemma2}
\end{align}}

where \(\cone\triangleq \p{\sqrt{\frac{1+q}{\np} + \frac{q}{\np}\bg^2 + \frac{\N-\np}{\np(\N-1)}\ghetsq}+\bg}^2\). Then,
\allowdisplaybreaks{\begin{align}
    & \E\normsq{\wtp-\w} \leq \E\normsq{\projwm\p{\wt-\beta\yxtrnd{1}^\top\yxtrnd{2}\wt}-\projwm\w} \nn \\
    & \leqos{\\\text{Contraction}\\\text{property of}\\\text{projection}} \E\normsq{\wt-\w-\beta\yxtrnd{1}^\top\yxtrnd{2}\wt} \nn \\
    & = \E\normsq{\wt-\w} - 2\beta\E\inp{\wt-\w}{\yxtrnd{1}^\top\yxtrnd{2}\wt} + \beta^2\E\normsq{\yxtrnd{1}^\top\yxtrnd{2}\wt} \nn \\
    & \eqos{\\\text{Using the in-}\\\text{dependence}\\\text{of }\rnd_1\text{ and }\rnd_2} \normsq{\wt-\w} - 2\beta\inp{\wt-\w}{\Jxt^\top\Jxt\wt} + \beta^2\E\normsq{\yxtrnd{1}^\top\yxtrnd{2}\wt} \nn \\
    % & = \normsq{\wt-\w} - 2\beta\inp{ \p{\wt-\w}}{\Jxt^\top\Jxt\wt} + \beta^2\E\normsq{\yxtrnd{1}^\top\yxtrnd{2}\wt} \nn \\
    & = \normsq{\wt-\w} - 2\beta\inp{ \Jxt\p{\wt-\w}}{\Jxt\wt} + \beta^2\E\normsq{\yxtrnd{1}^\top\yxtrnd{2}\wt} \nn \\
    &\leqos{\\ \Cref{line0:lemma2}} \normsq{\wt-\w} - 2\beta\inp{ \Jxt\p{\wt-\w}}{\Jxt\wt} + \beta^2\M\cone^2. \nn
\end{align}}
Then, arranging the terms, we get
\allowdisplaybreaks{\begin{align}
\inp{ \Jxt\p{\wt-\w}}{\Jxt\wt}\leq\frac{\E\normsq{\wt-\w}-\E\normsq{\wtp-\w}}{2\beta}+\frac{\beta\M\cone^2}{2}. \nn
\end{align}}

\paragraph{Proof of \Cref{lemma3}.}
\allowdisplaybreaks{\begin{align}
    &\E\normsq{\frac{1}{\n}\sum_{\ii\in\B^\tp{t}}\Dit} = \E\sqbr{\E\sqbr{\normsq{\frac{1}{\n}\sum_{\ii\in\B^\tp{t}}\Dit}\mid\wtp}} \nn \\
    &= \E\sqbr{\E\sqbr{\normsq{\frac{1}{\n}\sum_{\ii\in\B^\tp{t}} \frac{1}{\locit} \sum_{r=0}^{\locit-1} \sum_{\kk\in[\M]} \wktp \sG\fik\p{\x_\ii^\tp{t,k}}  }\mid\wtp}} \nn \\
    &\leqos{\\\text{Jensen's}\\\text{inequality}} \E\sqbr{\E\sqbr{\sum_{\kk\in[\M]} \wktp  \normsq{\frac{1}{\n}\sum_{\ii\in\B^\tp{t}} \frac{1}{\locit} \sum_{r=0}^{\locit-1} \sG\fik\p{\x_\ii^\tp{t,k}}  }\mid\wtp}} \nn \\
    & = \E\sqbr{\E\sqbr{\sum_{\kk\in[\M]} \wktp  \normsq{\frac{1}{\n}\sum_{\ii\in\B^\tp{t}} \frac{1}{\locit} \sum_{r=0}^{\locit-1} \sG\fik\p{\x_\ii^\tp{t,k}} \mp \G\fik\p{\x_\ii^\tp{t,k}}  }\mid\wtp}} \nn \\
    & \leqos{\\\text{Independence}\\\text{of SGD and}\\\text{Assump.}~\ref{assum:lochet}} \E\sqbr{\E\sqbr{\sum_{\kk\in[\M]} \wktp  \normsq{\frac{1}{\n}\sum_{\ii\in\B^\tp{t}} \frac{1}{\locit} \sum_{r=0}^{\locit-1} \G\fik\p{\x_\ii^\tp{t,k}}  }\mid\wtp}} + \E\sqbr{\E\sqbr{\sum_{\kk\in[\M]} \wktp  \frac{\lhetsq}{\n\locit}\mid\wtp}} \nn \\
    & \leqos{\\\text{Jensen's}\\\text{inequality}} \E\sqbr{\sum_{\kk\in[\M]} \wktp  \frac{1}{\n}\sum_{\ii\in\B^\tp{t}} \frac{1}{\locit} \sum_{r=0}^{\locit-1}\normsq{ \G\fik\p{\x_\ii^\tp{t,k}}  }} + \frac{\lhetsq}{\n\locit} \nn \\
    & \leqos{\\\text{Assump.~\ref{assum:bounded_grad}}} \E\sqbr{\sum_{\kk\in[\M]} \wktp  \frac{1}{\n}\sum_{\ii\in\B^\tp{t}} \frac{1}{\locit} \sum_{r=0}^{\locit-1}\bg^2 } + \frac{\lhetsq}{\n\locit} = \bg^2 + + \frac{\lhetsq}{\n\locit}. \nn
\end{align}}

\paragraph{Proof of \Cref{lemma4}.}
    \allowdisplaybreaks{\begin{align}
        &\E\sqbr{\inp{\Jxt\w}{-\p{\frac{1}{\n}\sum_{\ii\in\B^\tp{t}} \Dit -\Jxt\wtp}}} \nn \\
        & = \E\sqbr{ \E\sqbr{\inp{\Jxt\w}{-\p{\frac{1}{\n}\sum_{\ii\in\B^\tp{t}} \Dit -\Jxt\wtp}} \mid \wtp }  } \\
        & \eqos{\\\text{Expectation}\\\text{over SGD}\\\text{and }\B^\tp{t}} \E\sqbr{ \E\sqbr{\inp{\Jxt\w}{-\p{\frac{1}{\N}\sum_{\ii\in[\N]} \frac{1}{\locit}\sum_{r=0}^{\locit-1} \sum_{\kk\in[\M]}\wktp\G\fik\p{\x^\tp{t,r}} -\Jxt\wtp}} \mid \wtp } } \nn \\
        &\leqos{\\\text{Cauchy-Schwarz}}\E\sqbr{ \norm{\Jxt\w}\E\sqbr{ \norm{\frac{1}{N}\sum_{\ii\in[\N]}\frac{1}{\locit}\sum_{r=0}^{\locit-1} \sum_{\kk\in[\M]}\wktp\p{\G\fik\p{\x^\tp{t,r}} -\G \tl_\kk\pxt} }  \mid \wtp } } \nn \\
        &\eqos{\\\tl_\kk=\tfrac{1}{\N}\sum_\ii\fik}\E\sqbr{ \norm{\Jxt\w}\E\sqbr{ \norm{\frac{1}{N}\sum_{\ii\in[\N]}\frac{1}{\locit}\sum_{r=0}^{\locit-1} \sum_{\kk\in[\M]}\wktp\p{\G\fik\p{\x^\tp{t,r}} -\G\fik\pxt} }  \mid \wtp } } \nn \\
        & \leqos{\\\text{Assump.~\ref{assum:bounded_grad}}\\\text{Lemma~\ref{lemma0}}} \bg\E\sqbr{ \E\sqbr{ \norm{\frac{1}{N}\sum_{\ii\in[\N]}\frac{1}{\locit}\sum_{r=0}^{\locit-1} \sum_{\kk\in[\M]}\wktp\p{\G\fik\p{\x^\tp{t,r}} -\G\fik\pxt} }  \mid \wtp }  } \nn \\
        & \leqos{\\\text{Triangular}\\\text{inequality}} \bg\E\sqbr{ \E\sqbr{ \frac{1}{N}\sum_{\ii\in[\N]} \sum_{\kk\in[\M]}\wktp\norm{\frac{1}{\locit}\sum_{r=0}^{\locit-1} (\G\fik\p{\x^\tp{t,r}} -\G\fik\pxt) }  \mid \wtp }  } \nn \\
        & \leq \bg\E\sqbr{  \frac{1}{N}\sum_{\ii\in[\N]} \sum_{\kk\in[\M]}\wktp\E\sqbr{ \sqrt{\normsq{\frac{1}{\locit}\sum_{r=0}^{\locit-1} (\G\fik\p{\x^\tp{t,r}} -\G\fik\pxt) }}  \mid \wtp }  }\nn \\
        & \leqos{\\\text{Jensen's}\\\text{inequality}} \bg\E\sqbr{  \frac{1}{N}\sum_{\ii\in[\N]} \sum_{\kk\in[\M]}\wktp \sqrt{ \E\sqbr{\normsq{\frac{1}{\locit}\sum_{r=0}^{\locit-1} (\G\fik\p{\x^\tp{t,r}} -\G\fik\pxt) } \mid \wtp } }    }\nn \\
        & \leqos{\\\text{Lemma 2 in}\\\text{\cite{fedast}}} \bg\E\sqbr{  \frac{1}{N}\sum_{\ii\in[\N]} \sum_{\kk\in[\M]}\wktp \sqrt{ \frac{\smo^2\lrl^2\locit}{2(1-\smo^2\lrl^2\locit(\locit-1))}\lhetsq +\frac{\smo^2\lrl^2\locit(\locit-1))}{1-\smo^2\lrl^2\locit(\locit-1))}\normsq{\G\fik\pxt} }    } \nn \\
        & \leqos{\\\text{Assump.~\ref{assum:bounded_grad}}} \bg\E\sqbr{  \frac{1}{N}\sum_{\ii\in[\N]} \sum_{\kk\in[\M]}\wktp \sqrt{ \frac{\smo^2\lrl^2\locit}{2(1-\smo^2\lrl^2\locit(\locit-1))}\lhetsq +\frac{\smo^2\lrl^2\locit(\locit-1)}{1-\smo^2\lrl^2\locit(\locit-1)}\bg^2 }    } \nn \\
        & = \bg \sqrt{ \frac{\smo^2\lrl^2\locit}{2(1-\smo^2\lrl^2\locit(\locit-1))}\lhetsq +\frac{\smo^2\lrl^2\locit(\locit-1)}{1-\smo^2\lrl^2\locit(\locit-1)}\bg^2 } \nn \\
        & \leqos{\\\lrl\leq\mfrac{1}{2\smo\locit}\\\text{(assumption}\\\text{of the lemma)}} \bg \sqrt{ {\smo^2\lrl^2\locit}\lhetsq +{2\smo^2\lrl^2\locit(\locit-1)}\bg^2 } \leq \smo\lrl\sqrt{\locit}\lhet\bg + \smo\lrl\bg^2\sqrt{2\locit(\locit-1)}. \nn
    \end{align}}

\end{document}